\documentclass[10pt,twocolumn,letterpaper]{article}

\usepackage{iccv}
\usepackage{times}
\usepackage{epsfig}
\usepackage{graphicx}
\usepackage{amsmath}
\usepackage{amssymb}

\usepackage{tabularx}
\usepackage{booktabs}
\usepackage{caption}
\usepackage{makecell}
\usepackage{pifont}
\usepackage{multirow}

\usepackage{xcolor}

\newcommand{\cmark}{\textcolor{green}{\ding{51}}}%
\newcommand{\xmark}{\textcolor{red}{\ding{53}}}%

\newcolumntype{Y}{>{\centering\arraybackslash}X}

\newcommand{\Eq}[1]{Eq.~(\ref{eq:#1})}
\newcommand{\eq}[1]{\Eq{#1}}
\newcommand{\fig}[1]{Fig.~\ref{fig:#1}}
\newcommand{\tab}[1]{Table~\ref{tab:#1}}

\usepackage[pagebackref=true,breaklinks=true,letterpaper=true,colorlinks,bookmarks=false]{hyperref}

\iccvfinalcopy 

\def\iccvPaperID{8606} 

\ificcvfinal\fi

\begin{document}

\title{Zero-Shot Day-Night Domain Adaptation with a Physics Prior}

\author{Attila Lengyel\textsuperscript{1} \quad Sourav Garg\textsuperscript{2} \quad Michael Milford\textsuperscript{2} \quad Jan C. van Gemert\textsuperscript{1}\\
Delft University of Technology\textsuperscript{1} \quad QUT Centre for Robotics\textsuperscript{2}\\
{\tt\small \{a.lengyel, j.c.vangemert\}@tudelft.nl} \quad {\tt\small \{s.garg, michael.milford\}@qut.edu.au}
}

\maketitle

\begin{abstract}
  We explore the zero-shot setting for day-night domain adaptation. The traditional domain adaptation setting is to train on one domain and adapt to the target domain by exploiting unlabeled data samples from the test set. As gathering relevant test data is expensive and sometimes even impossible, we remove any reliance on test data imagery and instead exploit a visual inductive prior derived from physics-based reflection models for domain adaptation. We cast a number of color invariant edge detectors as trainable layers in a convolutional neural network and evaluate their  robustness to illumination changes. We show that the color invariant layer reduces the day-night distribution shift in feature map activations throughout the network. We demonstrate improved performance for zero-shot day to night domain adaptation on both synthetic as well as natural datasets in various tasks, including classification, segmentation and place recognition.
\end{abstract}


\section{Introduction}




Deep image recognition methods are sensitive to illumination shifts caused by accidental recording conditions such as camera viewpoint, light color, and illumination changes caused by time of day or weather~\cite{Afifi2019WhatEC,Dai18,Wulfmeier17}, as for example a model trained with daylight will not generalize to night time. Robustness to such recording conditions is essential for autonomous driving and other safety-critical computer vision applications. An illumination shift between train and test data is typically addressed by unsupervised domain adaptation~\cite{Romera2019BridgingTD,SDV20,Wang18} where the labeled training set is from one domain and the test set is from a different domain. The main assumption is that the test data is readily available and the challenge is how to make use of the unlabeled test data in an unsupervised setting to address the domain shift. However, adding test data is often non-trivial as it may be expensive and time consuming to obtain and due to the long tail of the real world is impossible to collect for all possible scenarios in advance.

Instead of adding more data, prior knowledge can be built-in as a visual inductive bias. The champion of such a bias is the convolution operator added to a deep network which yields a Convolutional Neural Network (CNN). The CNN is translation invariant, and thus saves a massive amount of data as the deep network no longer needs training samples at all possible locations. Here, we replace data by an inductive photometric bias. We introduce a novel zero-shot domain adaptation method for addressing day-night domain shifts exploiting learnable photometric invariant features as a physics-based visual inductive prior. In contrast to unsupervised domain adaptation, our zero-shot method reduces the data dependency by removing any reliance on the availability of test data.

{\renewcommand{\arraystretch}{0}%
\begin{figure}
\centering
\includegraphics[width=\linewidth]{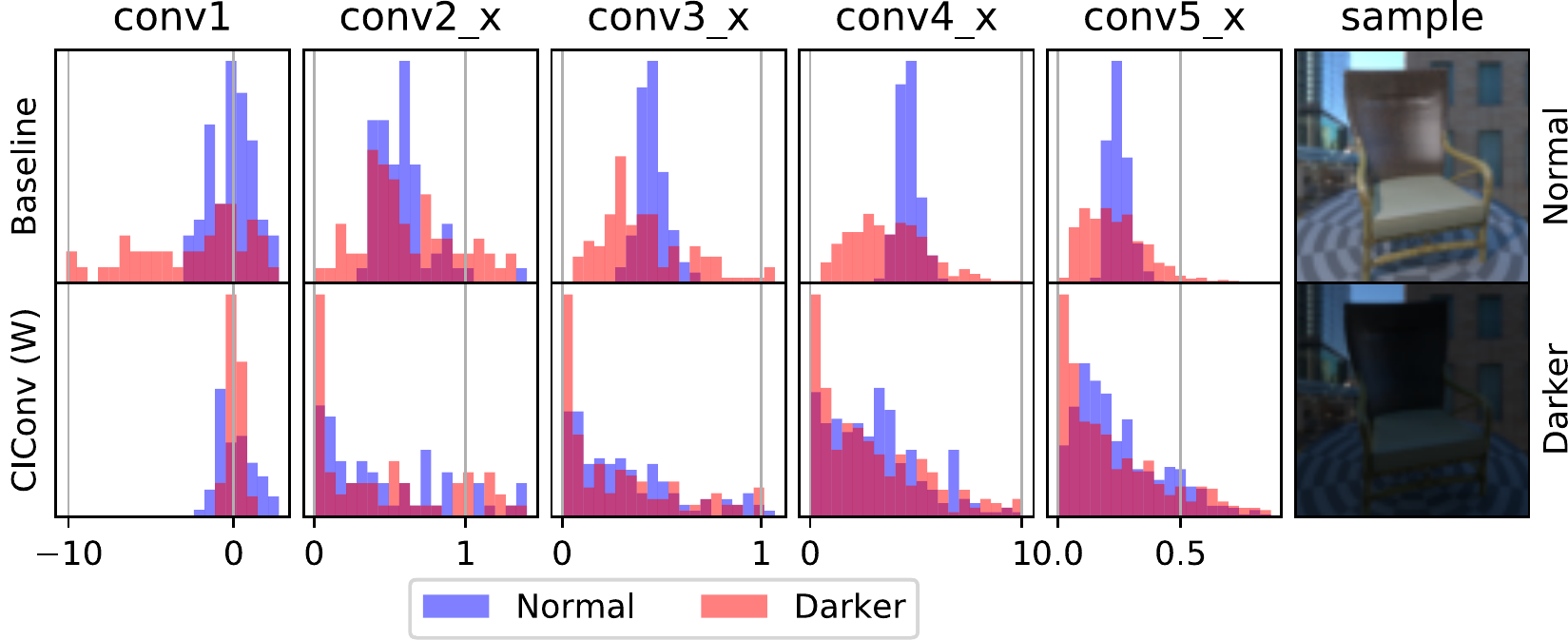}
\caption{Feature map activations in various layers of a baseline ResNet-18 and a color invariant $W$-ResNet-18, averaged over all samples in a `Normal' and `Darker' test set (samples on right). The intensity change between the test sets causes an internal distribution shift throughout all layers of the baseline model. $W$ normalizes the input resulting in more domain invariant features.}
\label{fig:activations}
\end{figure}}

Illumination changes to the source domain induce a distribution shift of feature map activations throughout all layers of a CNN. This is shown as the baseline in the top row of \fig{activations}, where the activations of a CNN trained on daytime data are shown for a `Normal' (source) and `Darker' (target) test set. Such a distribution shift, in turn, has a severe detrimental effect on the accuracy of the CNN~\cite{Li16}. Because the distribution shift is between the training data and unavailable test data, this  shift cannot be addressed in a data-driven manner using, for example, variants of Batch Normalization~\cite{IoffeS15,Li16}. Instead,  we normalize feature map activations in a data-free setting by exploiting photometric invariant features which are explicitly designed to tackle distribution shifts caused by illumination changes.

Photometric invariant features, or color invariants, represent object properties irrespective of the accidental recording conditions~\cite{Geusebroek2001Color,gevers1999color}, including 1) scene geometry, which affects the formation of shadows and shading, the 2) color and 3) intensity of the light source, which changes the overall tint and brightness of the scene, and 4) Fresnel reflections occurring on shiny materials where the incoming light is directly reflected from the surface without interacting with the material color. Thanks to their robustness to these lighting changes, color invariants have been widely used in classical computer vision applications \cite{Alvarez11,Maddern14}, yet their use in a deep learning setting has remained largely unexplored. We implement the color invariant edge detectors from~\cite{Geusebroek2001Color} as a trainable Color Invariant Convolution (CIConv) layer which can be used as the input layer to any CNN to transform the input to a domain invariant representation.   \fig{activations}, bottom row, shows that CIConv reduces the distribution shift  between the source and target test set in all network layers, improving target domain performance.
    
We have the following contributions: (i) we introduce CIConv, a learnable color invariant CNN layer that reduces the activation distribution shift in a CNN under an illumination-based domain shift; (ii) we evaluate several color invariants in the day-night domain adaptation setting on our two carefully curated classification datasets; and (iii) we demonstrate performance improvements on tasks related to autonomous driving, including classification, segmentation and place recognition. All datasets and code will be made available on our project page.\footnote{\url{https://github.com/Attila94/CIConv}}
\section{Related work}

\paragraph{Domain Adaptation} 
The aim of domain adaptation~\cite{Wang18} is to train a model on a source domain dataset such that it performs well on a different but similar target domain dataset. This alleviates the burden of annotating datasets for applications in new domains where insufficient training data is available. Popular approaches rely on generative adversarial networks (GANs) to generate synthetic target domain samples~\cite{Hoffman18} or aim to minimize the feature divergence between the two domains through an adversarial term~\cite{Hoffman16,Tzeng17} or a discrepancy metric~\cite{TzengHZSD14,Long15} in the loss function. The day-night domain adaptation setting is particularly important due to the promise of self-driving cars and thus includes much work~\cite{cho2020semantic,Dai18,di2020rainy,Romera2019BridgingTD,Sakaridis19,SDV20,Sun19,valada2017adaptive,vertens2020heatnet,Wulfmeier17} for semantic segmentation, and~\cite{Anoosheh2019NighttoDayIT,Jenicek2019NoFO,Porav18} for place recognition. However, all aforementioned methods (except~\cite{cho2020semantic}) require either training data from the target domain or additional modalities, whereas our approach uses only source domain image data. Our approach requires no extra information sources and thus preempts expensive data gathering costs.

\paragraph{Zero-shot Domain Adaptation}
Research on zero-shot learning~\cite{akata2015label,lampert2013attribute,mensink2014costa,norouzi2013zero,xian2018zeroShotEval,Zhang_2016_CVPR} has been readily extended from unseen classes to unseen domains, where domain adaptation is performed without having access to the target domain. However, current zero-shot domain adaptation methods require additional information in the form of: (i) extra task-irrelevant source and target domain data pairs to adapt to the task-relevant target domain~\cite{Peng2017ZeroShotDD,Wang_2019_ICCV}; (ii) a parametrization of the domain shift by an attribute, where the attribute probability distribution for the unseen target domain is required to be known~\cite{Ishii2019ZeroshotDA}; (iii) additional data from domains besides the source and target domain to learn a domain-invariant subspace projection~\cite{YangH15b}, or; (iv) extra data in a partially labelled target domain~\cite{Wang2019}. These four types of information are generally not known for day-night domain shifts and are therefore not directly applicable. AdaBN~\cite{Li16} argues that domain-specific knowledge is stored in the batch normalization (BN)~\cite{IoffeS15} layers of a model and performs domain adaptation by resampling BN statistics from the target domain. This again requires access to the target domain dataset. AdaBN~\cite{Li16} can be considered zero-shot if only the statistics of the current batch are used. However, this makes the method reliant on large batch sizes where classes are evenly represented. In contrast, our method does not require any information from the target domain other than the task agnostic physics-based illumination prior given by color invariants which are readily available from literature.

\paragraph{Physics-Guided Neural Networks}
Adding prior knowledge from physical models in a neural network has the potential to improve performance without additional training data. The canonical example is adding translation equivariance through a convolutional prior~\cite{kayhanCVPR20spatialLocation,urbanICLR16CNNdeepAndC} where recent work shows benefits from adding prior knowledge, for example in line detection~\cite{lin2020deepHoughPrior}, spectral leakage~\cite{tomen2021spectralLeakHammingPrior} and anti-aliasing in CNNs~\cite{Zhang2019}.
In the case of physical image formation models, recent examples include intrinsic image decomposition~\cite{Baslamisli17}, underwater image enhancement~\cite{Zhou2020DomainAA}, or rain image restoration~\cite{Li19rain}. Here, we add an physical image formation prior to compensate for the lack of data in zero-shot domain adaptation. We investigate a relatively unexplored direction combining deep learning with physical color and reflection invariants.

\paragraph{Color invariants}
The use of physics-based reflection models to improve invariance to illumination changes is a well-researched topic in classical computer vision~\cite{barnard1997color,burghouts2009performance,funt1995color,gevers1999color,vanDeSande2009evaluating,weijer2005boosting,weijer005edge}. Early work includes invariants derived from the Kubelka-Munk (KM) reflection model~\cite{Kubelka99,Geusebroek2001Color}. Based on the image formation model introduced in~\cite{Finlayson01} various methods have been proposed for shadow removal or intrinsic image decomposition~\cite{Finlayson09,Finlayson06} with applications in place recognition~\cite{Corke13,Maddern14}, road detection~\cite{Alvarez08,Alvarez11,Kim17,Krajnik15} and street image segmentation~\cite{Upcroft14}.  Recent works have shown improved segmentation performance by applying a color invariant transformation as a preprocessing step~\cite{Alshammari18,Alshammari2019MultiTaskLF,Maxwell2019RealTimePR} or using the ground truth albedo as input on a synthetic dataset~\cite{BaslamisliECCV2018}. \cite{Afifi2019WhatEC} demonstrates the sensitivity of CNNs to changes in white balance (WB) settings and shows how robustness can be improved using an auto-WB preprocessing step. Our work further explores the use of classical color invariants as a trainable deep network layer.

\section{Method}
\label{sec.method}

Our color invariant layers make use of the invariant edge detectors from~\cite{Geusebroek2001Color}. The edge detectors are derived from the image formation model based on the Kubelka-Munk theory~\cite{Kubelka99} for material reflections, which describes the spectrum of light $E$ reflected from an object in the viewing direction as
\begin{align}
\label{eq:kubelka_munk}
 E(\lambda,\mathbf{x}) = e(\lambda,\mathbf{x})\left((1-\rho_f(\mathbf{x}))^2R_\infty(\lambda,\mathbf{x})+\rho_f(\mathbf{x})\right)
\end{align}
where $\bf x$ denotes the spatial location on the image plane, $\lambda$ the wavelength of the light, $e(\lambda,\mathbf{x})$ the spectrum of the light source, $R_\infty$ the material reflectivity and $\rho_f$ the Fresnel reflectance coefficient. Partial derivatives of $E$ with respect to $x$ and $\lambda$ are denoted by subscripts $E_x$ and $E_\lambda$, respectively.

A color invariant representation does not rely on accidental scene properties such as lighting and viewing direction and depends only on the material property $R_\infty$. By exploring simplifying assumptions in \eq{kubelka_munk}, we can derive various invariant representations, as summarized in~\tab{color_invariants}. The derived invariants $E$, $W$, $C$, $N$ and $H$ represent edge detectors that are invariant to various combinations of illumination changes, including scene geometry (i.e. does not detect shadow and shading edges), Fresnel reflections, and the intensity and color of the illuminant. For the complete derivations of the color invariants in \tab{color_invariants}, we refer to Section \textcolor{red}{1} of the supplementary material.

\begin{table*}[ht]
    \centering
    \begin{tabularx}{\textwidth}{@{}YlYYYY@{}}
        \toprule
        \textbf{Invariant} & \multicolumn{1}{c}{\textbf{Definition}} & \textbf{SG} & \textbf{FR} & \textbf{II} & \textbf{IC} \\ \midrule
        $E$ &
        $E = \sqrt{E_x^2 + E_{\lambda x}^2 + E_{\lambda \lambda x}^2 + E_y^2 + E_{\lambda y}^2 + E_{\lambda \lambda y}^2}$ &
        \xmark & \xmark & \xmark & \xmark \\ \midrule
        $W$ &
        \makecell[l]{$W = \sqrt{W_x^2 + W_{\lambda x}^2 + W_{\lambda \lambda x}^2 + W_y^2 + W_{\lambda y}^2 + W_{\lambda \lambda y}^2},$\\
        $W_x=\frac{E_x}{E}, \quad W_{\lambda x}=\frac{E_{\lambda x}}{E}, \quad W_{\lambda\lambda x}=\frac{E_{\lambda\lambda x}}{E}$} &
        \xmark & \xmark & \cmark & \xmark \\ \midrule
        $C$ &
        \makecell[l]{$C = \sqrt{C_{\lambda x}^2 + C_{\lambda \lambda x}^2 + C_{\lambda y}^2 + C_{\lambda \lambda y}^2},$\\
        $C_{\lambda x}=\frac{E_{\lambda x}E-E_{\lambda}E_x}{E^2}, \quad C_{\lambda\lambda x}=\frac{E_{\lambda\lambda x}E-E_{\lambda\lambda}E_x}{E^2}$} &
        \cmark & \xmark & \cmark & \xmark \\ \midrule
        $N$ &
        \makecell[l]{$N = \sqrt{N_{\lambda x}^2 + N_{\lambda \lambda x}^2 + N_{\lambda y}^2 + N_{\lambda \lambda y}^2},$\\
        $N_{\lambda x}=\frac{E_{\lambda x}E-E_{\lambda}E_x}{E^2}, N_{\lambda\lambda x}=\frac{E_{\lambda\lambda x}E^2-E_{\lambda\lambda}E_xE-2E_{\lambda x}E_{\lambda}E+2E_{\lambda}^2E_x}{E^3}$} &
        \cmark & \xmark & \cmark & \cmark \\ \midrule
        $H$ &
        \makecell[l]{$H = \sqrt{H_x^2 + H_y^2}, \quad H_x=\frac{E_{\lambda\lambda}E_{\lambda x}-E_{\lambda}E_{\lambda\lambda x}}{E_{\lambda}^2+E_{\lambda\lambda}^2}$} &
        \cmark & \cmark & \cmark & \xmark \\
        \bottomrule
    \end{tabularx}
    \caption{Overview of color invariant edge detectors~\cite{Geusebroek2001Color} and their invariance properties to \textbf{S}cene \textbf{G}eometry, \textbf{F}resnel \textbf{R}eflections, \textbf{I}llumination \textbf{I}ntensity, \textbf{I}llumination \textbf{C}olor. $E$ is a baseline intensity edge detector and is not invariant to any changes. Subscripts denote partial derivatives, where $\lambda$ is the spectral derivative and  $x$ the spatial derivative of \eq{kubelka_munk}. Spatial derivatives for the $y$ direction follow directly from the ones given for the $x$ direction.}
    \label{tab:color_invariants}
\end{table*}

The Gaussian color model \cite{Geusebroek2001Color} is used to estimate $E$, $E_\lambda$ and $E_{\lambda\lambda}$ from the RGB camera responses as
\begin{align}
\label{eq.rgb_gaussian}
  \begin{bmatrix}
  E(x,y)\\
  E_\lambda(x,y)\\
  E_{\lambda\lambda}(x,y)\\
  \end{bmatrix}
  \!=\!
  \begin{bmatrix}
  0.06 & 0.63 & 0.27\\
  0.3 & 0.04 & -0.35\\
  0.34 & -0.6 & 0.17
  \end{bmatrix}\!
  \begin{bmatrix}
  R(x,y)\\G(x,y)\\B(x,y)
  \end{bmatrix}
\end{align}
where $x,y$ are pixel location in the image.
Spatial derivatives $E_x$ and $E_y$ are calculated by convolving $E$ with a Gaussian derivative kernel $g$ with standard deviation $\sigma$, i.e.
\begin{align}
\label{eq:gaussder}
E_x(x,y,\sigma) &= \sum_{t \in \mathbb{Z}} E(t,y) \frac{\partial g(x-t,\sigma)}{\partial x}
\end{align}
and similarly for $E_y$, $E_{\lambda x}$, $E_{\lambda\lambda x}$, $E_{\lambda y}$ and $E_{\lambda\lambda y}$. Finally, the color invariant edge map is defined as the gradient magnitude of all relevant spatial derivatives as shown in \tab{color_invariants}.

\begin{figure}[]
    \setlength\tabcolsep{0pt}
    {\renewcommand{\arraystretch}{0}
    \begin{tabularx}{\linewidth}{YYY}
        \includegraphics[width=\linewidth]{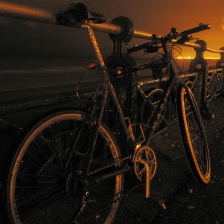} &
        \includegraphics[width=\linewidth]{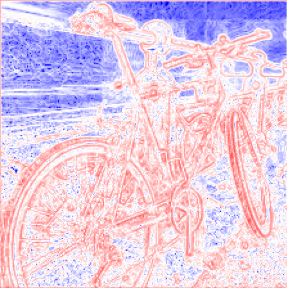} &
        \includegraphics[width=\linewidth]{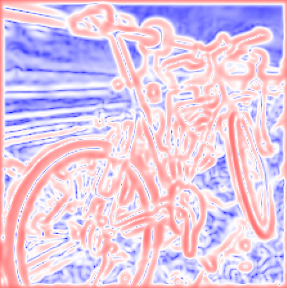} \\ \addlinespace[4pt]
        Input & $\sigma=0.50$ & $\sigma=2.00$
    \end{tabularx}}
    \caption{Color invariant representation $W$ of the input image for two different values of $\sigma$. Note the trade-off between detail (small $\sigma$) and noise robustness (large $\sigma$).}
    \label{fig:scale}
\end{figure}

The $\sigma$ parameter in \eq{gaussder} determines the scale at which the image is convolved with the Gaussian derivative filters and as such the amount of detail preserved in the color invariant representation of an image. A small $\sigma$ results in a detailed edge map but is more sensitive to noise, whereas a large $\sigma$ is more robust but may omit important details. A visualization is given in \fig{scale} for color invariant $W$. Rather than fixing $\sigma$ a-priori we implement the edge detector as a trainable layer to learn the task-specific optimal scale.  The resulting Color Invariant Convolution (CIConv) is used as the input layer of the CNN and outputs a single channel representation onto which subsequent convolutional layers can be stacked. For computational simplicity we omit the square root from the gradient magnitude of the color invariants, and apply a log transformation and sample-wise normalization such that the distribution of the edge maps is close to standard normal. Furthermore, instead of directly optimizing $\sigma$, we train a scale parameter $s$ such that $\sigma=2^s$. This stabilizes training by reducing the backpropagation gradient for small values of $s$ and ensures that $\sigma$ is always positive. CIConv is thus defined as
\begin{align}
\label{eq.nconv}
\text{CIConv}(x,y) = \frac{\log{ \left( \text{CI}^2(x,y,\sigma=2^s) + \epsilon \right)} - \mu_\mathcal{S} }{\sigma_\mathcal{S}}
\end{align}
with CI the color invariant of choice from Tab. \ref{tab:color_invariants}, $\mu_\mathcal{S}$ and $\sigma_\mathcal{S}$ the sample mean and standard deviation over $\log{\left(\text{CI}^2+\epsilon\right)}$, and $\epsilon$ a small term added for numerical stability.

\section{Experiments}


\subsection{Illumination robustness of CNNs}
\label{sec:shapenet}
\begin{figure*}[]
    \setlength\tabcolsep{0pt}
    \begin{tabularx}{\linewidth}{YYYYYYYYYY}
        \includegraphics[width=\linewidth]{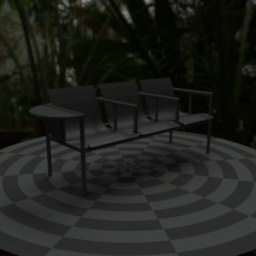} &
        \includegraphics[width=\linewidth]{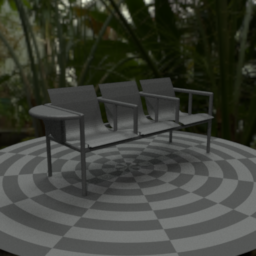} &
        \includegraphics[width=\linewidth]{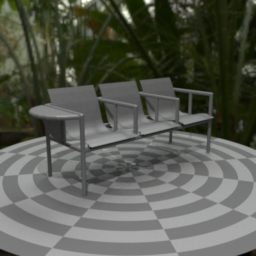} &
        \includegraphics[width=\linewidth]{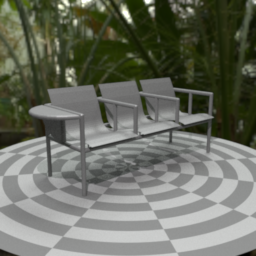} &
        \includegraphics[width=\linewidth]{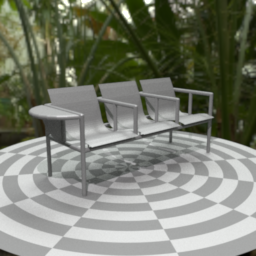} & 
        \includegraphics[width=\linewidth]{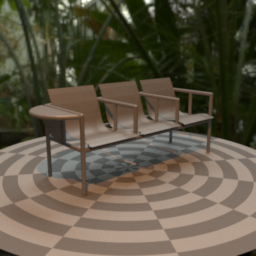} &
        \includegraphics[width=\linewidth]{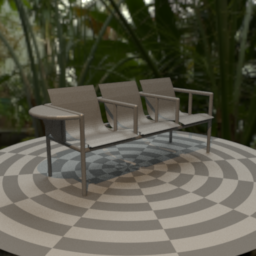} &
        \includegraphics[width=\linewidth]{figures/shapenet/000278_0_6500K.png} &
        \includegraphics[width=\linewidth]{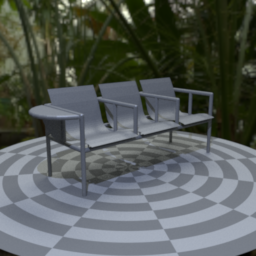} &
        \includegraphics[width=\linewidth]{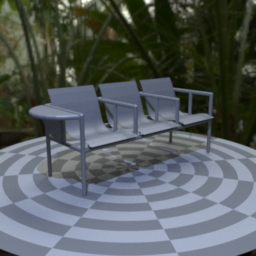} \\
        Darker & 
        Dark & 
        Normal & 
        Light & 
        Lighter &
        2500K &
        4000K & 
        6500K & 
        12000K & 
        20000K
    \end{tabularx}
    \caption{Sample from the synthetic classification dataset rendered from ShapeNet~\cite{shapenet2015}, shown in all illumination conditions represented in the test set. The five leftmost samples correspond to a varying light source intensity, whereas in the five rightmost samples a range of light source temperatures is shown. ``Normal'' and ``6500K'' are equivalent.}
    \label{fig:shapenet-illuminants}
\end{figure*}

We investigate to what degree CIConv improves a CNN's robustness to accidental recording conditions by performing a classification experiment on a synthetic image dataset where we have accurate control over the illumination of the scene. The images are rendered from a subset of the ShapeNet~\cite{shapenet2015} dataset using the physically based renderer Mitsuba~\cite{Mitsuba}. The scene is illuminated by a point light modeled as a black-body radiator with temperatures ranging between $[1900,20000]K$ and an ambient light source. The training set contains 1,000 samples for each of the 10 object classes recorded under ``normal'' lighting conditions ($T=6500K$). Multiple test sets with 300 samples per class are rendered for a variety of light source intensities and colors. \fig{shapenet-illuminants} shows an overview of the illumination conditions represented in the test set.

\paragraph{CIConv improves illumination robustness} We train a baseline ResNet-18~\cite{he16deep} and five models with the CIConv layer with invariants $E$, $W$, $C$, $N$ and $H$, respectively. Training is done for 175 epochs with a batch size of 64 using SGD with momentum 0.9, weight decay 1e-4 and an initial learning rate of 0.05 with stepwise reduction by factor 0.1, step size 50. Data augmentation is performed in the form of random horizontal flips, random cropping and random rotations. The models are evaluated on both test sets and the average classification accuracy over three runs is shown in~\fig{shapenet-results}. The accuracy of the baseline RGB model quickly drops as lighting conditions start to diverge from the training set. The performance of the color invariant networks remains more stable with $W$ consistently outperforming all others.

{\renewcommand{\arraystretch}{0}%
\begin{figure}
\includegraphics[width=\linewidth]{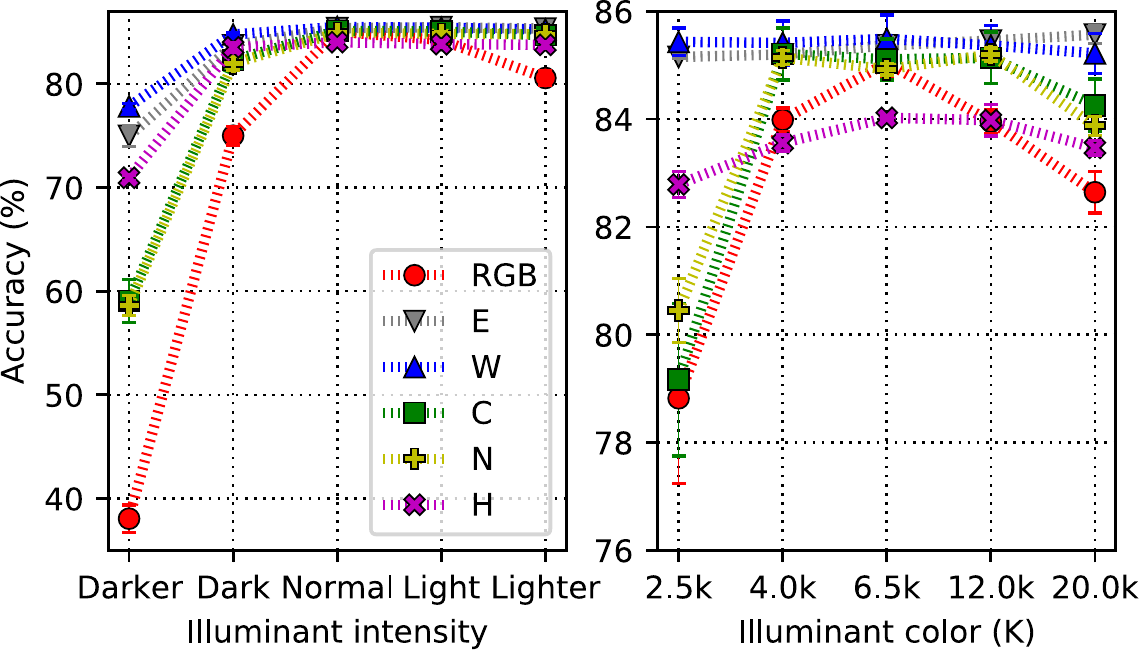}
\caption{Classification accuracy of ResNet-18 with various color invariants on the synthetic ShapeNet dataset. RGB (not invariant) performance degrades when illumination conditions differ between train and test set, while color invariants remain more stable. $W$ performs best overall.}
\label{fig:shapenet-results}
\end{figure}}

\paragraph{CIConv reduces feature map distribution shift} The robustness of the color invariant networks compared to the baseline can be explained by analyzing the feature map activations of the networks. We calculate the mean feature map activation in different layers of the networks, averaged over all samples in the Normal and Dark test sets. The histograms in \fig{activations} show that the intensity change between the normal and low light test sets caused a clear distribution shift throughout all network layers of the baseline model. In contrast, the CIConv layer with invariant $W$ produces a domain invariant feature representation and consequently the distributions in the network are more aligned between the two domains. We quantify the distribution shift as the L2 distance between feature maps for the two domains, where again $W$ yields the smallest distance. The L2 distances as well as histograms of the distributions of feature map activations for other color invariants are provided in section \textcolor{red}{2} of the supplementary material.


\subsection{Day-night natural image classification}

\begin{figure*}[]
    \setlength\tabcolsep{0pt}
    {\renewcommand{\arraystretch}{0}
    \begin{tabularx}{\linewidth}{YYYYYYYYYY}
        \includegraphics[width=\linewidth]{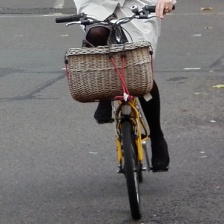} &
        \includegraphics[width=\linewidth]{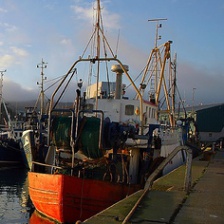} &
        \includegraphics[width=\linewidth]{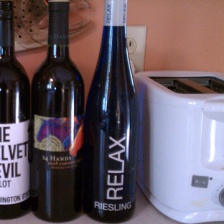} &
        \includegraphics[width=\linewidth]{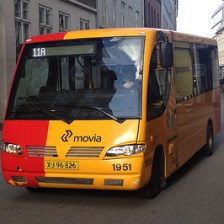} &
        \includegraphics[width=\linewidth]{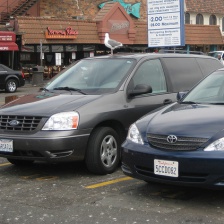} &
        \includegraphics[width=\linewidth]{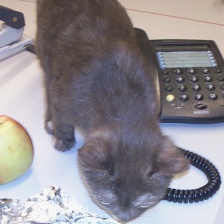} &
        \includegraphics[width=\linewidth]{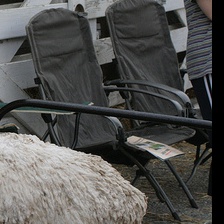} &
        \includegraphics[width=\linewidth]{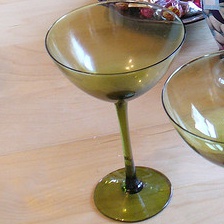} &
        \includegraphics[width=\linewidth]{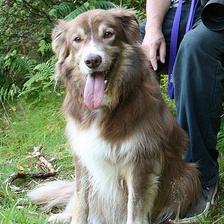} &
        \includegraphics[width=\linewidth]{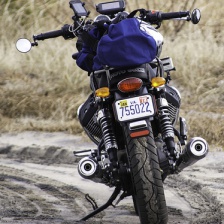} \\ 
        \includegraphics[width=\linewidth]{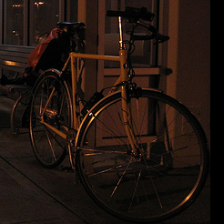} &
        \includegraphics[width=\linewidth]{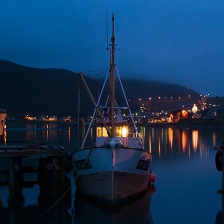} &
        \includegraphics[width=\linewidth]{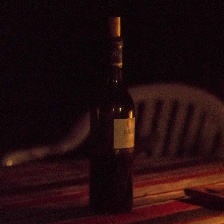} &
        \includegraphics[width=\linewidth]{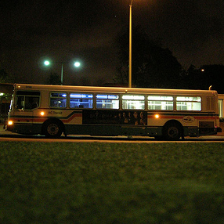} &
        \includegraphics[width=\linewidth]{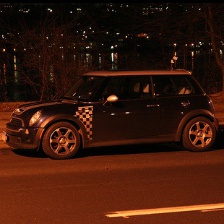} &
        \includegraphics[width=\linewidth]{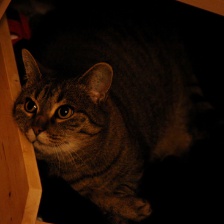} &
        \includegraphics[width=\linewidth]{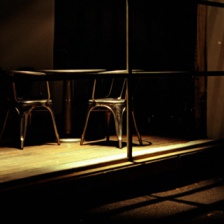} &
        \includegraphics[width=\linewidth]{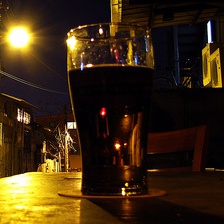} &
        \includegraphics[width=\linewidth]{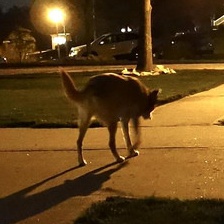} &
        \includegraphics[width=\linewidth]{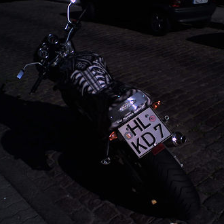} \\ \addlinespace[4pt]
        Bicycle & 
        Boat & 
        Bottle & 
        Bus & 
        Car &
        Cat &
        Chair & 
        Cup & 
        Dog & 
        Motorbike
    \end{tabularx}}
    \caption{Samples from the day (source domain) and night (target domain) test sets of the CODaN dataset.}
    \label{fig:codan}
\end{figure*}

To verify that the properties of the color invariants also generalize to natural images we perform a classification experiment on a novel day-to-night dataset. We present the Common Objects Day and Night (CODaN) dataset, consisting of images from 10 common object classes recorded in both day and nighttime. It contains a daytime training set of 1,000 samples per class, a daytime validation set of 50 samples per class, and separate day and night test sets of 300 samples per class. CODaN is composed from the ImageNet~\cite{imagenet_cvpr09}, COCO~\cite{mscoco14} and ExDark~\cite{loh2019getting} datasets. Samples of the day and night test sets are shown in \fig{codan}.

\paragraph{Performance on natural images} We trained color invariant versions of ResNet-18 on CODaN using the same settings as in \ref{sec:shapenet}, but without random cropping and with random brightness, contrast, hue and saturation augmentations. \tab{codan_results} shows the accuracy of the baseline and the color invariant networks, averaged over three runs. Additionally, other color invariants (luminance, normalized RGB, comprehensive normalization~\cite{finlayson98comprehensive} and others~\cite{Alvarez11,Maddern14}) are evaluated, which are implemented as a preprocessing step. We also consider a slightly adjusted version of AdaBN as a possible zero-shot domain adaptation method, which provides a significant performance increase by sampling the batch statistics for the Batch Normalization layers during test time for each individual batch. This is opposed to the original AdaBN method, where the batch statistics are calculated from the target domain dataset a priori. $W$ outperforms all other models on the nighttime test set by a large margin. The luminance baseline performs surprisingly well, whereas the other non-trainable color invariants even result in a performance drop.

\begin{table}[t]
\begin{tabularx}{\linewidth}{@{}l YY@{}}
\toprule
\textbf{Method} & \textbf{Day} & \textbf{Night}  \\ \midrule
Baseline & 80.39 $\pm$ 0.38 & 48.31 $\pm$ 1.33 \\
$E$ & 79.79 $\pm$ 0.40 & 49.95 $\pm$ 1.60 \\
$W$ & \textbf{81.49} $\pm$ \textbf{0.49} & \textbf{59.67} $\pm$ \textbf{0.93} \\
$C$ & 78.04 $\pm$ 1.08 & 53.44 $\pm$ 1.28 \\
$N$ & 77.44 $\pm$ 0.00 & 52.03 $\pm$ 0.27 \\
$H$ & 75.20 $\pm$ 0.56 & 50.52 $\pm$ 1.34 \\
Luminance & 80.67 $\pm$ 0.32 & 51.37 $\pm$ 0.58 \\
Normalized RGB & 63.44 $\pm$ 1.52 & 41.66 $\pm$ 1.56 \\
Comprehensive norm.~\cite{finlayson98comprehensive} & 70.52 $\pm$ 1.10 & 44.34 $\pm$ 1.57 \\
Alvarez and Lopez~\cite{Alvarez11} & 64.41 $\pm$ 0.74 & 30.06 $\pm$ 0.57 \\
Maddern et al.~\cite{Maddern14} & 60.83 $\pm$ 0.98 & 33.04 $\pm$ 1.28 \\
AdaBN~\cite{Li16} & 79.72 $\pm$ 0.59 & 55.55 $\pm$ 1.07 \\ \midrule
\textbf{Ablations} & \textbf{Day} & \textbf{Night}  \\ \midrule
Baseline + norm. & 63.43 $\pm$ 1.32 & 42.15 $\pm$ 0.98 \\
Baseline + log + norm. & 63.49 $\pm$ 0.55 & 41.90 $\pm$ 0.69 \\
Baseline w/o color aug. & 78.99 $\pm$ 0.59 & 36.00 $\pm$ 0.59 \\
$W$ w/o color aug. & 79.71 $\pm$ 0.57 & 53.62 $\pm$ 0.88 \\
\bottomrule
\end{tabularx}
\caption{CODaN classification accuracy of a ResNet-18 architecture with various color invariants (top). $W$ performs best.
Ablation studies (bottom) show the individual effect of normalization, log scaling and photometric augmentations.}
\label{tab:codan_results}
\end{table}


\paragraph{Color invariant transformations on natural images}
We visualize the $E$, $W$, $C$, $N$ and $H$ color invariant transformations of a day and night test sample (RGB) in \fig{ciconv}. $E$ being a non-invariant edge detector has low edge strengths in low intensity parts of the dark image. $W$ on the other hand normalizes for intensity, yielding a more constant edge map. $C$, $N$ and $H$ are invariant to changes in scene geometry and therefore do not detect edges with low color saturation, resulting in significant information loss. In addition, these invariants seem to be more amplifying the noise in low intensity parts of the image. Overall, $W$ is able to 1) detect low intensity and low saturation edges and 2) suppress noise in low-intensity parts of the image, and therefore produces the most robust and informative edge map.

{\renewcommand{\arraystretch}{0}%
\begin{figure*}
\setlength\tabcolsep{0pt}
\begin{tabular}{cccccc}
RGB & $E$ & $W$ & $C$ & $N$ & $H$ \\ \addlinespace[4pt]
\includegraphics[width=0.165\linewidth]{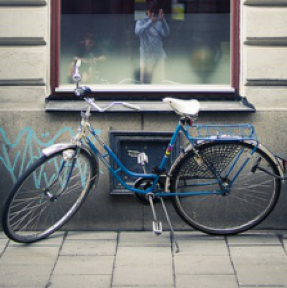} &
\includegraphics[width=0.165\linewidth]{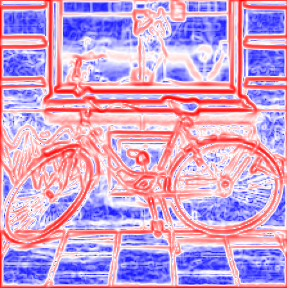} &
\includegraphics[width=0.165\linewidth]{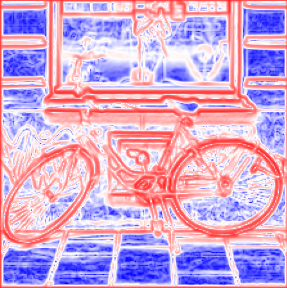} &
\includegraphics[width=0.165\linewidth]{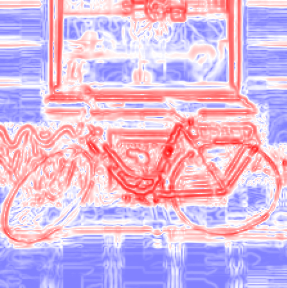} &
\includegraphics[width=0.165\linewidth]{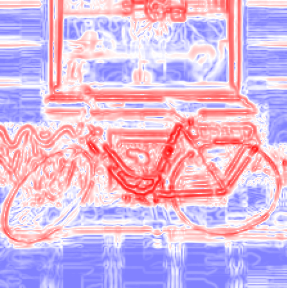} &
\includegraphics[width=0.165\linewidth]{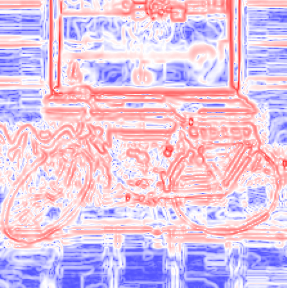} \\
\includegraphics[width=0.165\linewidth]{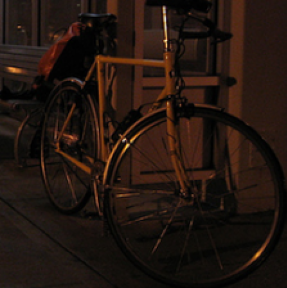} &
\includegraphics[width=0.165\linewidth]{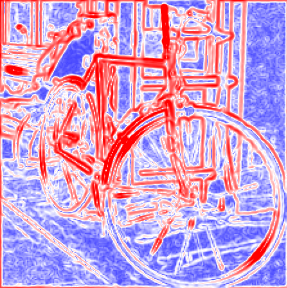} &
\includegraphics[width=0.165\linewidth]{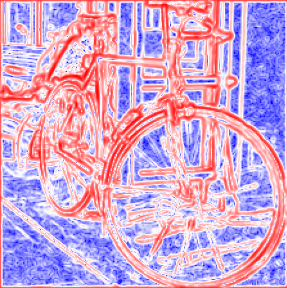} &
\includegraphics[width=0.165\linewidth]{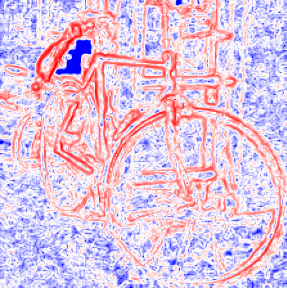} &
\includegraphics[width=0.165\linewidth]{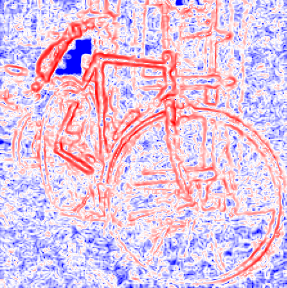} &
\includegraphics[width=0.165\linewidth]{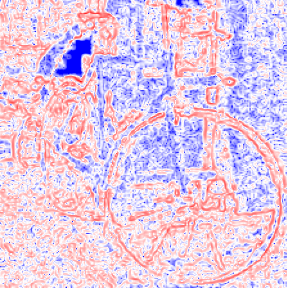} \\
\end{tabular}
\caption{Color invariant visualizations of day and night samples from CODaN (red: positive; blue: negative values). $E$ does not detect low intensity edges, whereas $C$, $N$ and $H$ do not detect edges that have low color saturation. $W$ produces the most robust and informative edge map.}
\label{fig:ciconv}
\end{figure*}}

\begin{figure}
\centering
\includegraphics[width=\columnwidth]{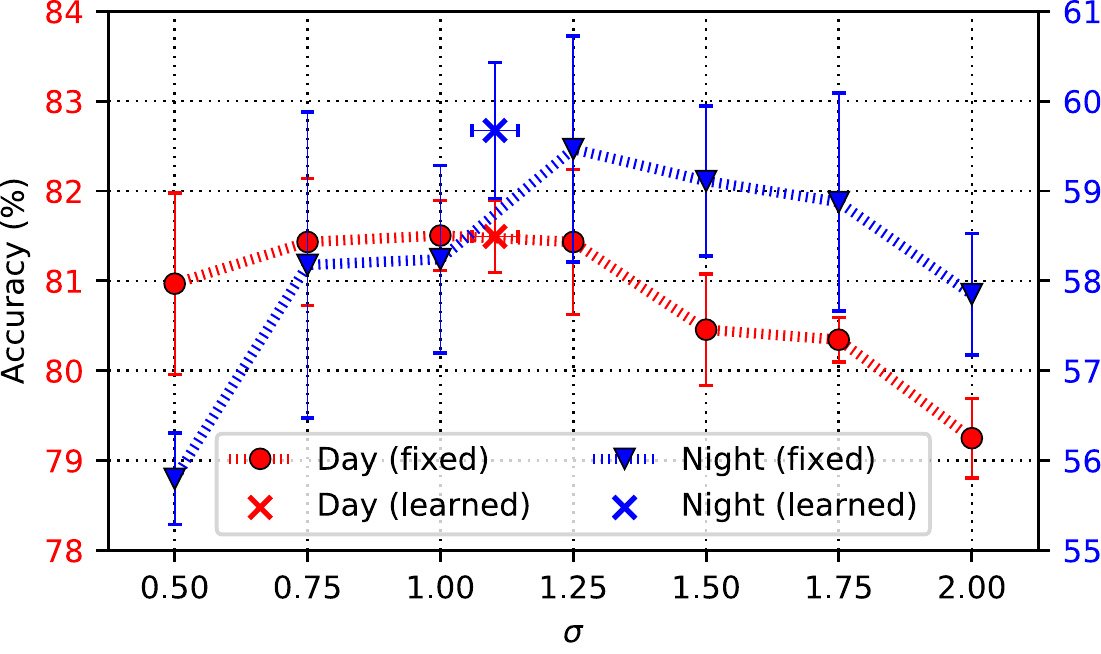}
\caption{Performance on CODaN day (left y-axis) and night (right y-axis) test sets for various fixed values of $\sigma$. Learned $\sigma$ and corresponding accuracies are indicated by crosses. CIConv learns the optimal value.}
\label{fig:effect_sigma}
\end{figure}

\paragraph{Learned vs. fixed scale} We verify that CIConv learns the optimal scale by training the model with a range of fixed $\sigma$ values, using invariant $W$. \fig{effect_sigma} shows the average accuracy over five runs. We observe that selecting the wrong scale $\sigma$ has a detrimental effect on accuracy. When the scale is learnable, it converges to the optimal value for the daytime dataset, as indicated by the red cross in the figure. This value proves also optimal for the nighttime domain.

\paragraph{Ablation studies}
We evaluate whether simple log scaling and sample-wise normalization of RGB images, without applying a color invariant transformation, can achieve the same improved performance on the nighttime test set. Furthermore, we investigate how the baseline and $W$ networks perform when trained without brightness, contrast, hue and saturation augmentations. The results are shown in the bottom part of \tab{codan_results}. Normalization, both with and without log scaling, does not yield better performance for the baseline model. This indicates that addressing the distribution shift between the source and target domain observed in the feature map activations of a network requires more than simple intensity normalization of the input sample. Moreover, photometric augmentations mostly seem to benefit the baseline network, whereas the model with color invariant $W$ is inherently more robust to illumination changes. Both results underscore the importance and effectiveness of the color invariant transformation.


\subsection{Semantic segmentation}
We perform a semantic segmentation experiment using the RefineNet~\cite{Lin2017RefineNet} architecture with ResNet-101 and $W$-ResNet-101 feature extractors pre-trained on the ImageNet~\cite{imagenet_cvpr09} dataset. The segmentation model is trained on the training set of the CityScapes~\cite{Cordts2016Cityscapes} dataset containing 2,975 densely annotated daytime street images and evaluated on the 50 coarsely annotated street images from Nighttime Driving~\cite{Dai18} and the 151 densely annotated images from the Dark Zurich~\cite{Sakaridis19} test set. We perform training using SGD with momentum 0.9, weight decay 1e-4 and an initial learning rate of 0.1 which is step-wise reduced by a factor 0.1 after every 30 epochs. All input images are resized to 1024x512 pixels and randomly cropped to 768x384 pixels, allowing a batch size of 6 on 2 GeForce GTX 1080 Ti GPUs. Data augmentation is applied by random scaling, brightness-, contrast- and hue-shifting, and horizontal flipping. Inference is done on 1024x512 samples without cropping.

Results are shown in \tab{segmentation_results} as the mean Intersection-over-Union (mIoU). Results for other methods are taken from their corresponding papers. The color invariant $W$-RefineNet significantly outperforms the vanilla RefineNet and RefineNet-AdaBN models, which are also trained only on source domain data, and has competitive performance compared to methods trained on both source and target domain data. Qualitative segmentation results are shown in \fig{segmentation}. Detailed per-class scores are included in section \textcolor{red}{4} of the supplementary material.

\begin{table}[]
\begin{tabularx}{\linewidth}{@{}l YY@{}}
\toprule
\textbf{Method} & \makecell{\textbf{Nighttime} \\ \textbf{Driving}} & \makecell{\textbf{Dark} \\ \textbf{Zurich}} \\ \midrule
\multicolumn{2}{@{}l}{\textbf{Trained on source data only}} \\ \midrule
RefineNet~\cite{Lin2017RefineNet} & 34.1 & 30.6 \\
$W$-RefineNet [ours] & \textbf{41.6} & \textbf{34.5} \\
RefineNet-AdaBN~\cite{Li16} & 36.3 & 31.3 \\ \midrule
\multicolumn{2}{@{}l}{\textbf{Trained on source and target data}} \\ \midrule
ADVENT~\cite{vu2018advent} & 34.7 & 29.7 \\
BDL~\cite{Li_2019_CVPR} & 34.7 & 30.8 \\
AdaptSegNet~\cite{tsai2018learning} & 34.5 & 30.4 \\
DMAda~\cite{Dai18} & 41.6 & 32.1 \\
Day2Night~\cite{Sun19} & 45.1 & - \\
GCMA~\cite{Sakaridis19} & 45.6 & 42.0   \\
MGCDA~\cite{SDV20} & \textbf{49.4} & \textbf{42.5} \\ \bottomrule
\end{tabularx}
\caption{Segmentation performance on Nighttime Driving~\cite{Dai18} and Dark Zurich~\cite{Sakaridis19}, reported as mIoU scores. $W$-RefineNet outperforms other methods trained only on daytime data and has competitive performance to methods also using nighttime images.}
\label{tab:segmentation_results}
\end{table}

\begin{figure*}[]
    \setlength\tabcolsep{0pt}
    {\renewcommand{\arraystretch}{0}\footnotesize
    \begin{tabularx}{\linewidth}{YYYYYYY}
        \includegraphics[width=\linewidth,height=0.5625\linewidth]{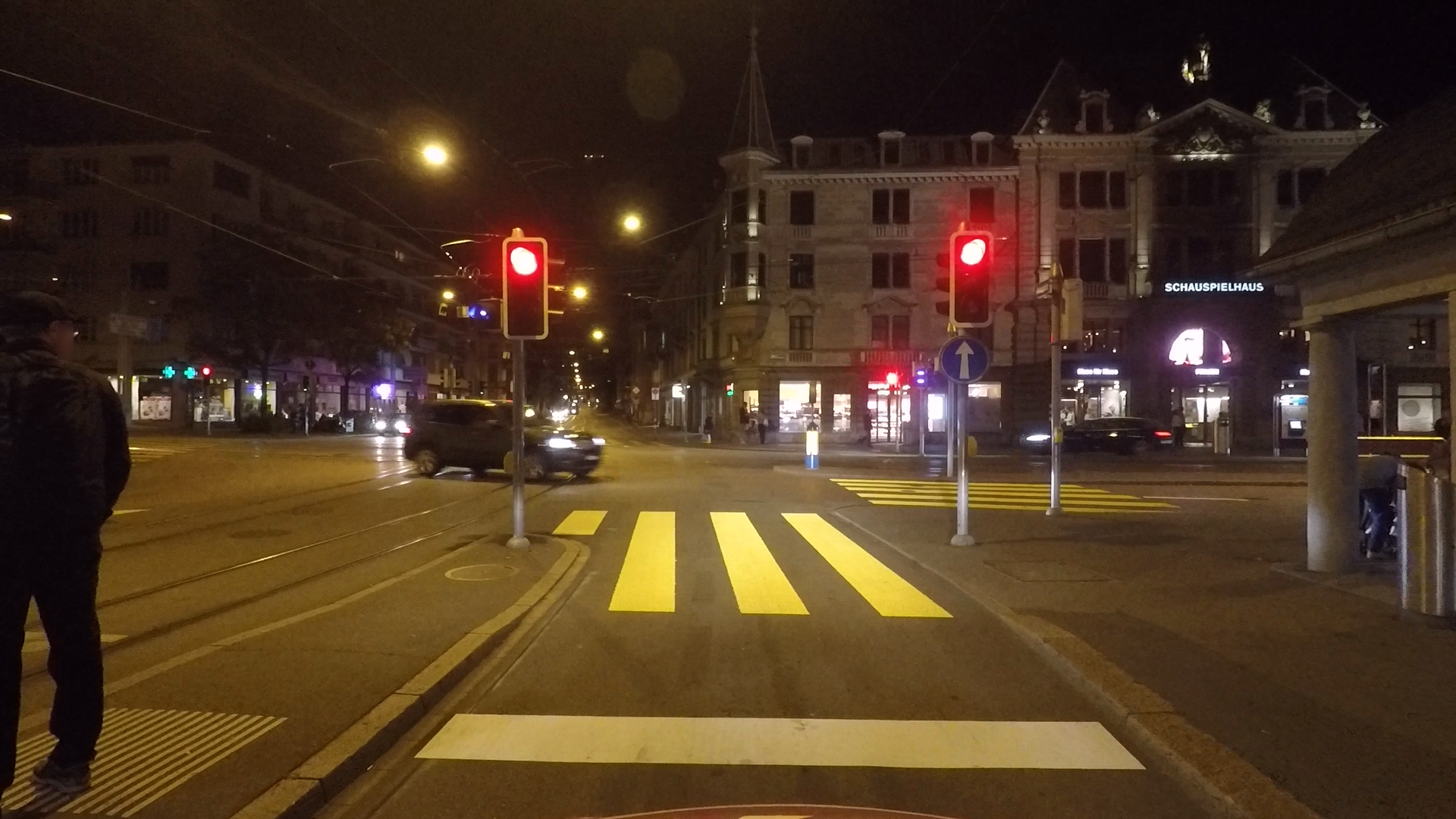} &
        \includegraphics[width=\linewidth,height=0.5625\linewidth]{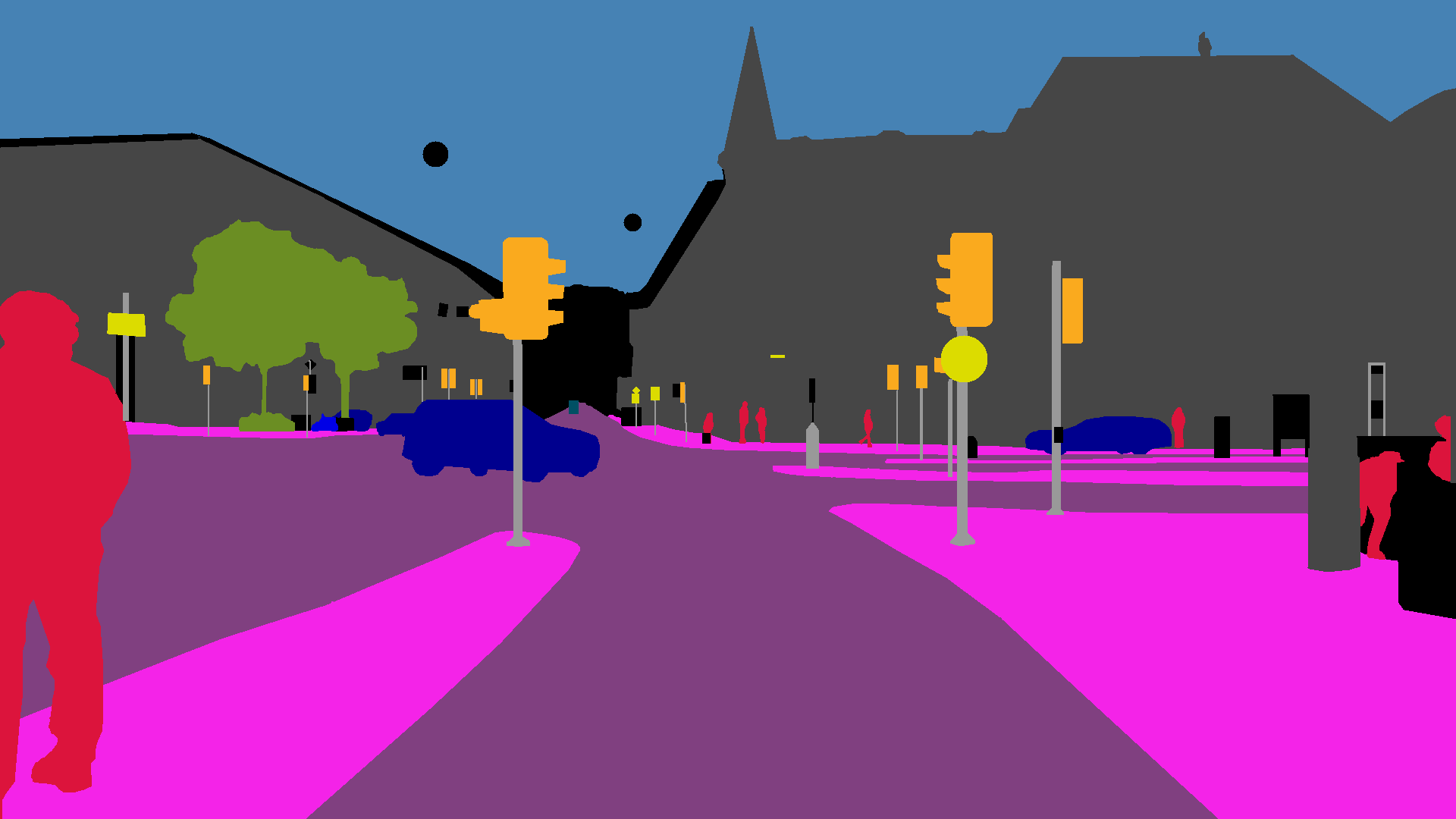} &
        \includegraphics[width=\linewidth,height=0.5625\linewidth]{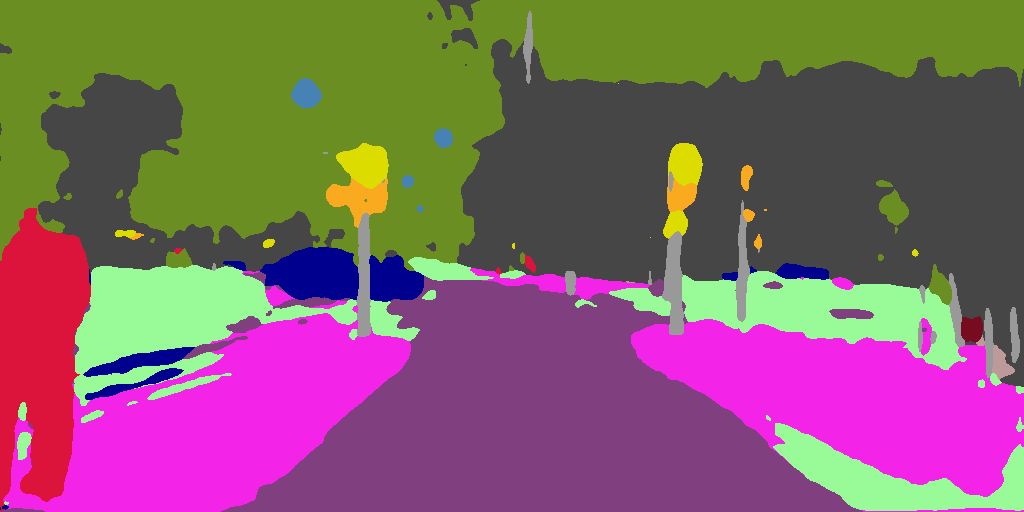} &
        \includegraphics[width=\linewidth,height=0.5625\linewidth]{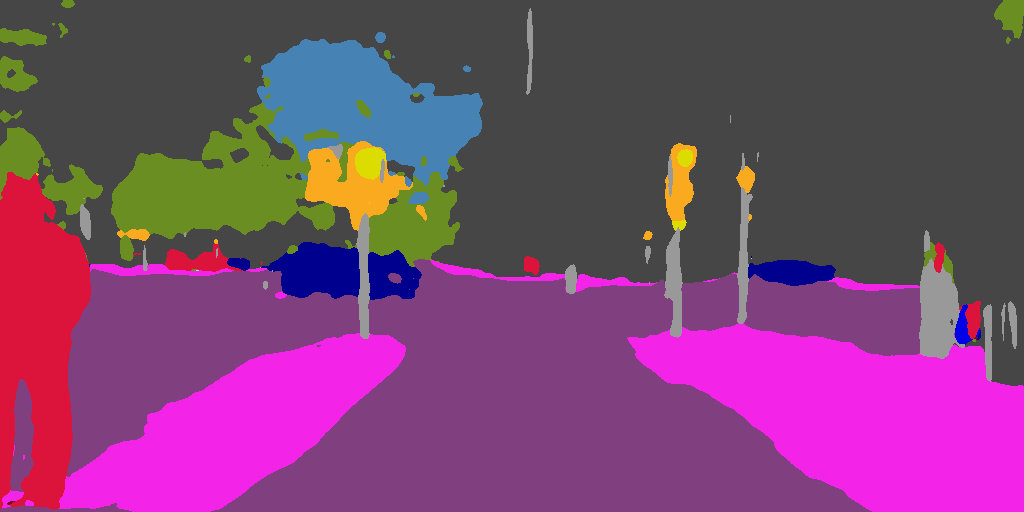} &
        \includegraphics[width=\linewidth,height=0.5625\linewidth]{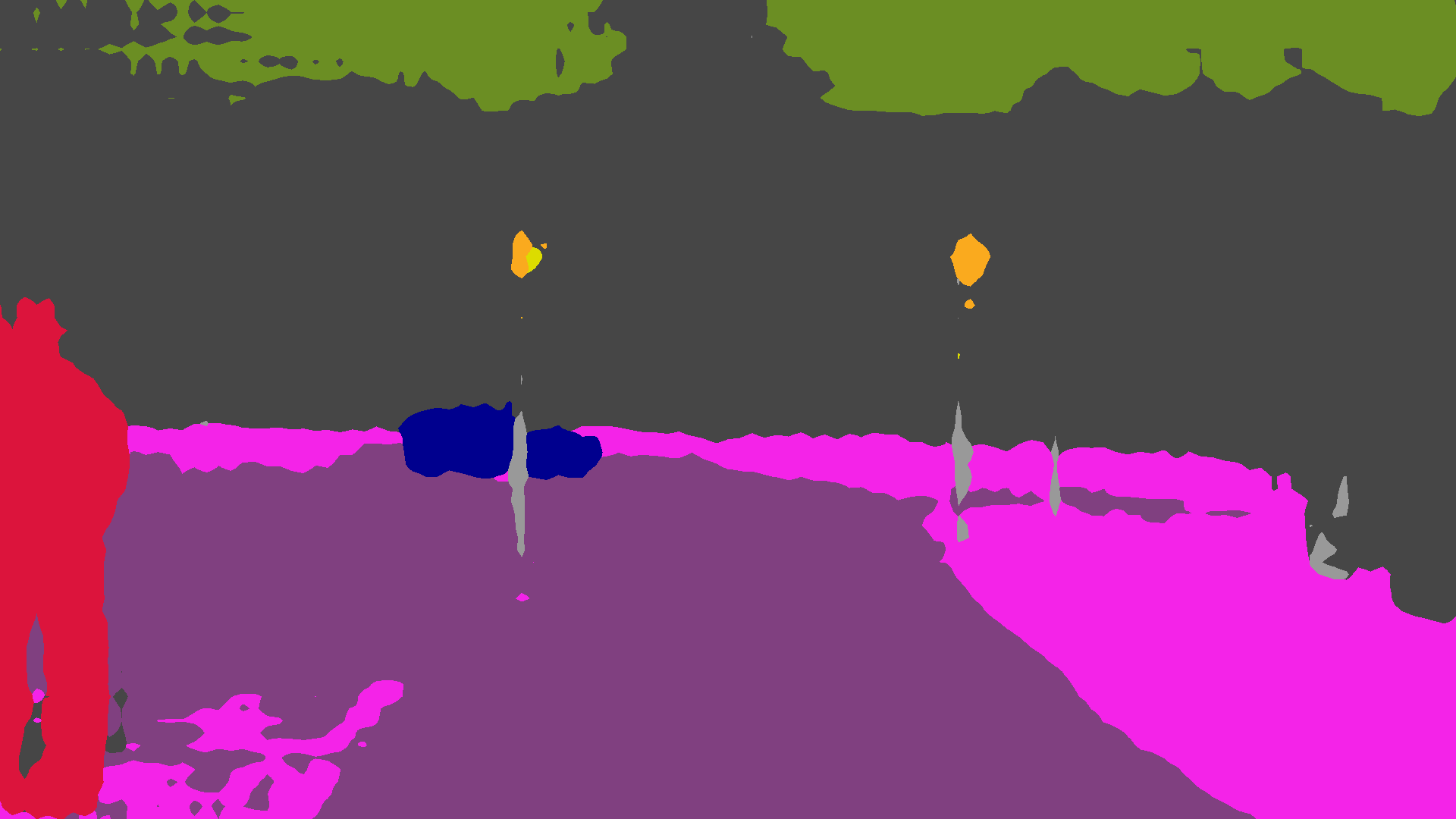} &
        \includegraphics[width=\linewidth,height=0.5625\linewidth]{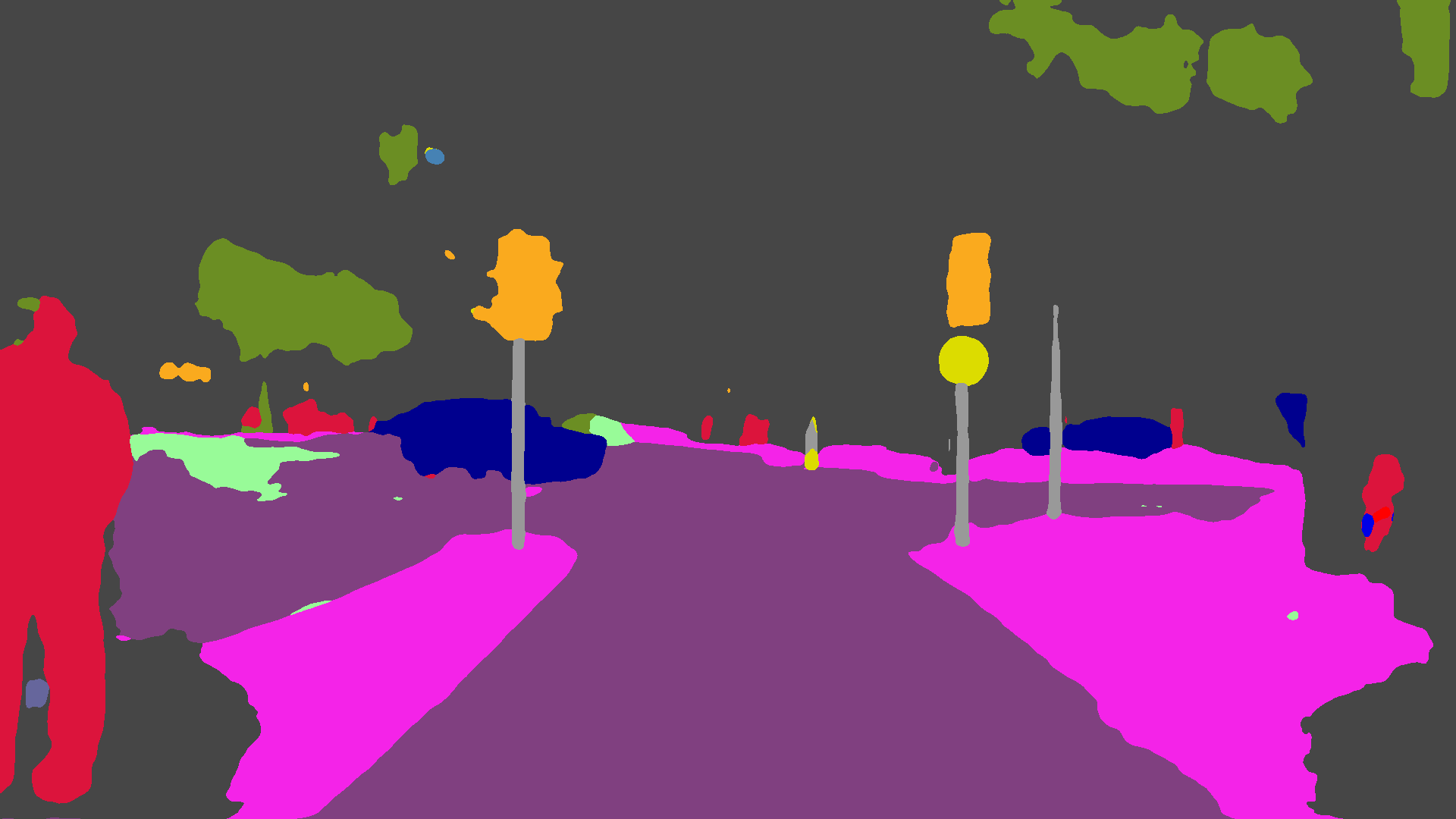} &
        \includegraphics[width=\linewidth,height=0.5625\linewidth]{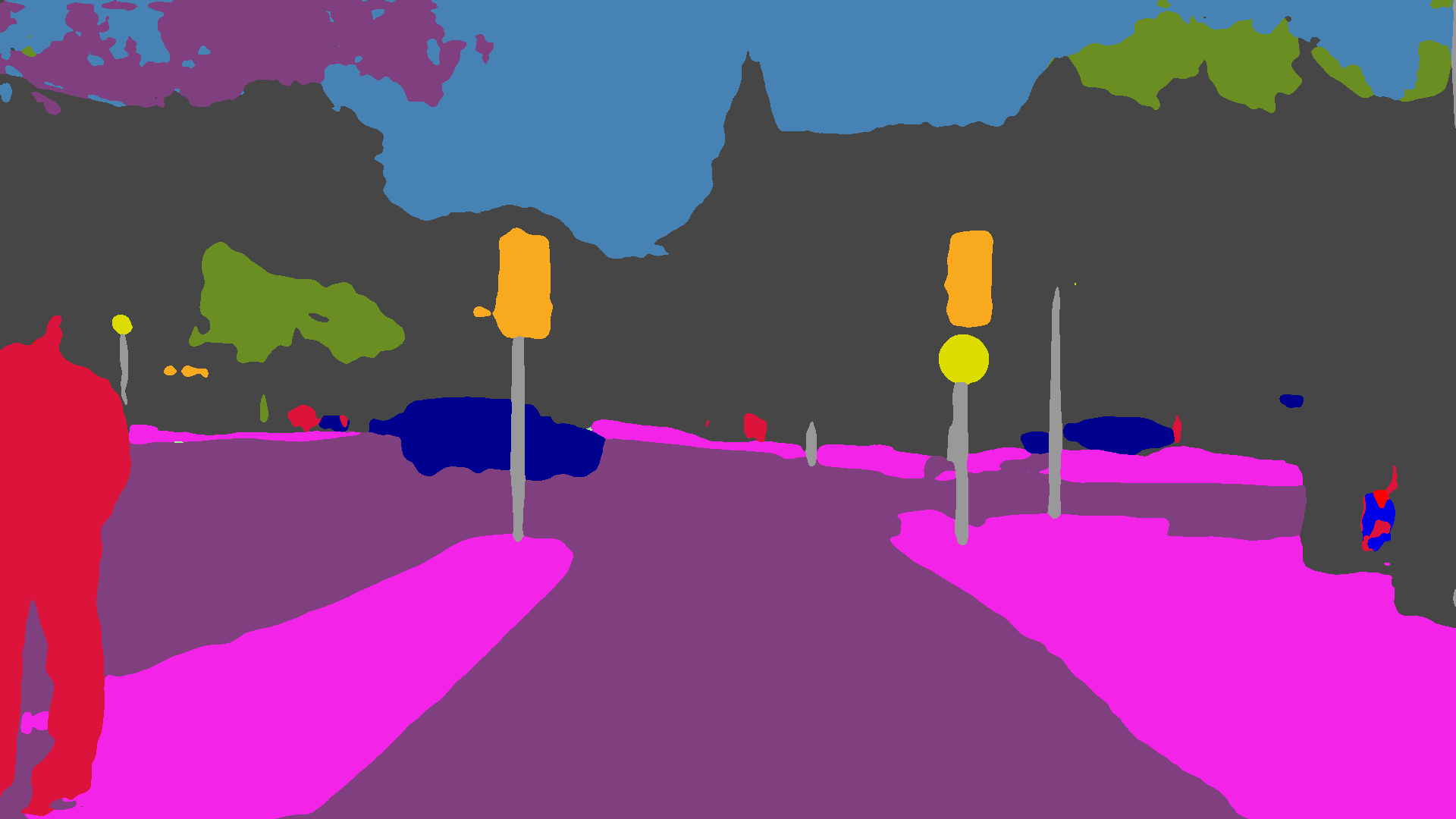} \\
        \includegraphics[width=\linewidth,height=0.5625\linewidth]{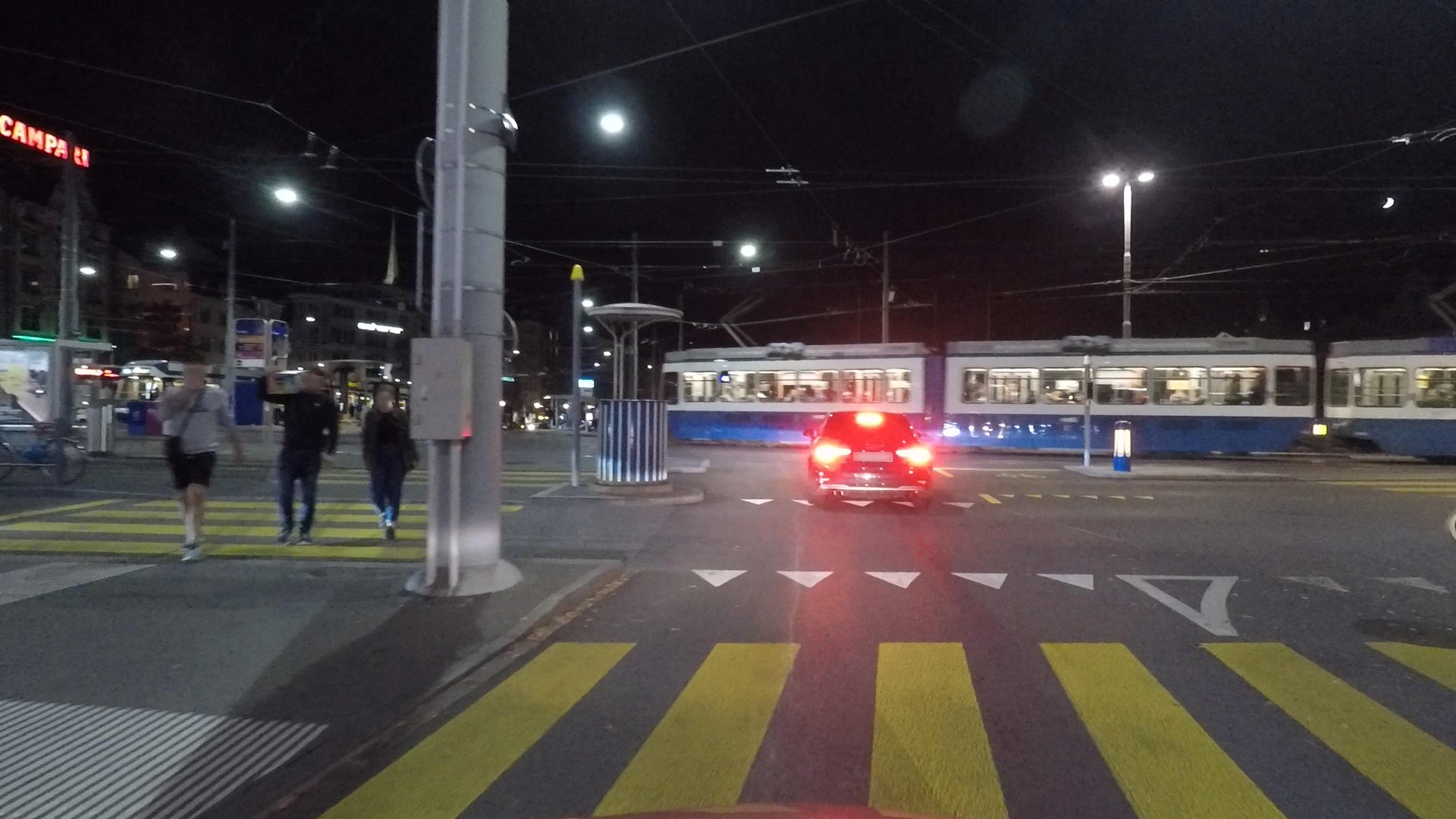} &
        \includegraphics[width=\linewidth,height=0.5625\linewidth]{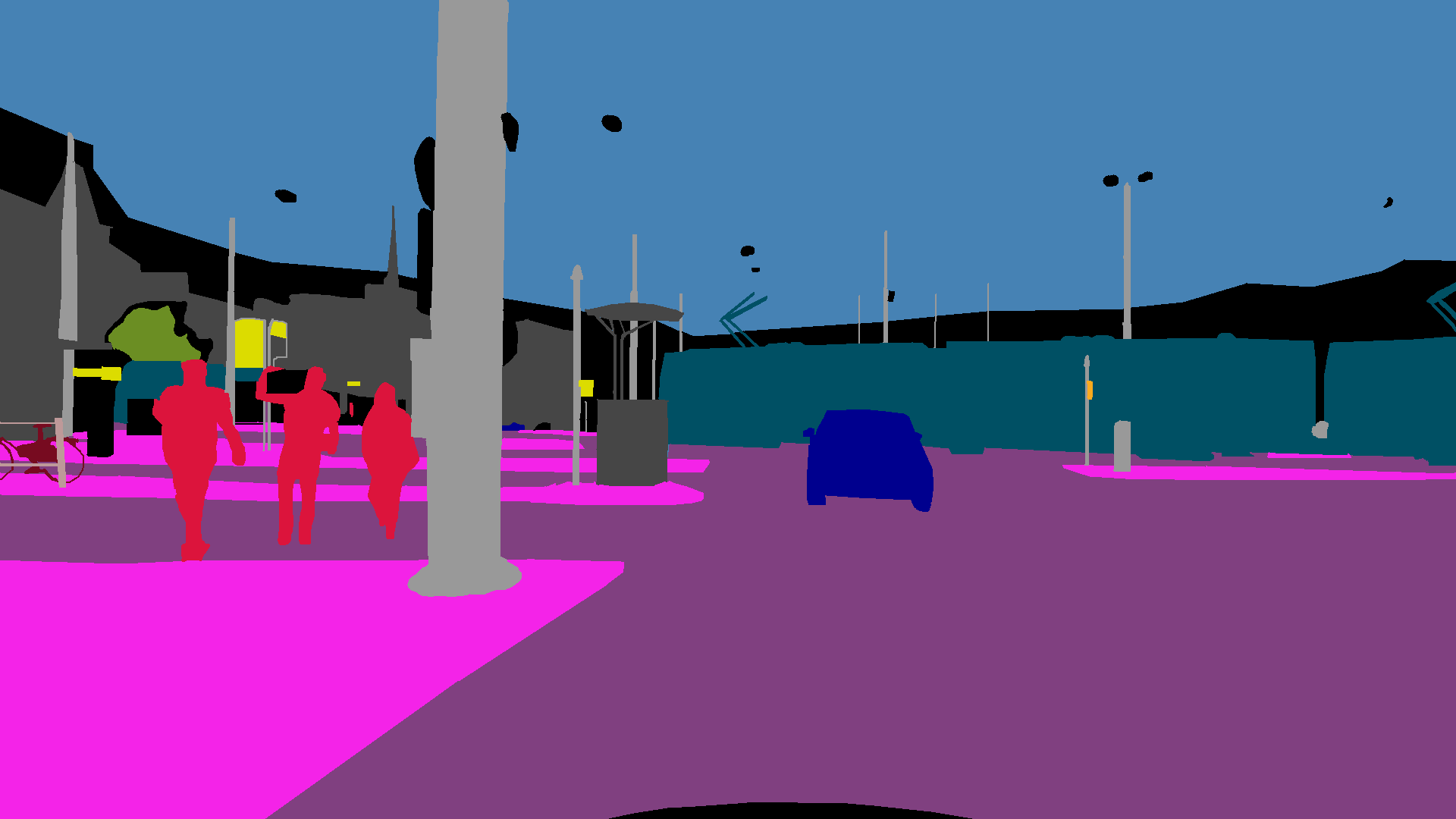} &
        \includegraphics[width=\linewidth,height=0.5625\linewidth]{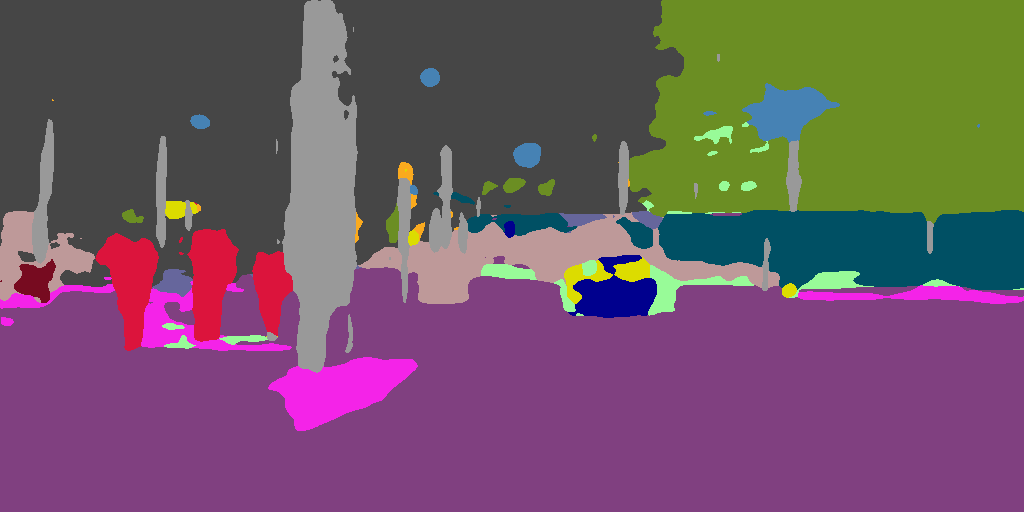} &
        \includegraphics[width=\linewidth,height=0.5625\linewidth]{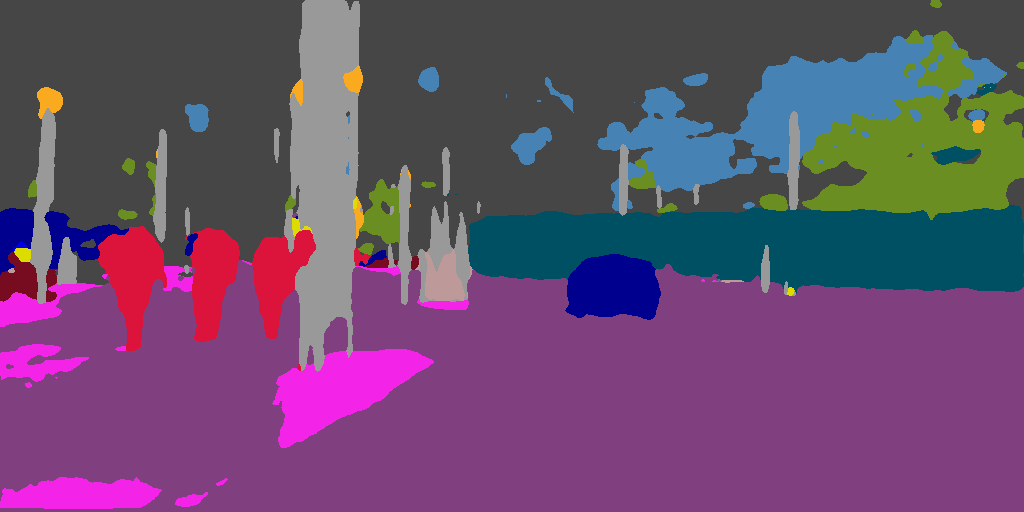} &
        \includegraphics[width=\linewidth,height=0.5625\linewidth]{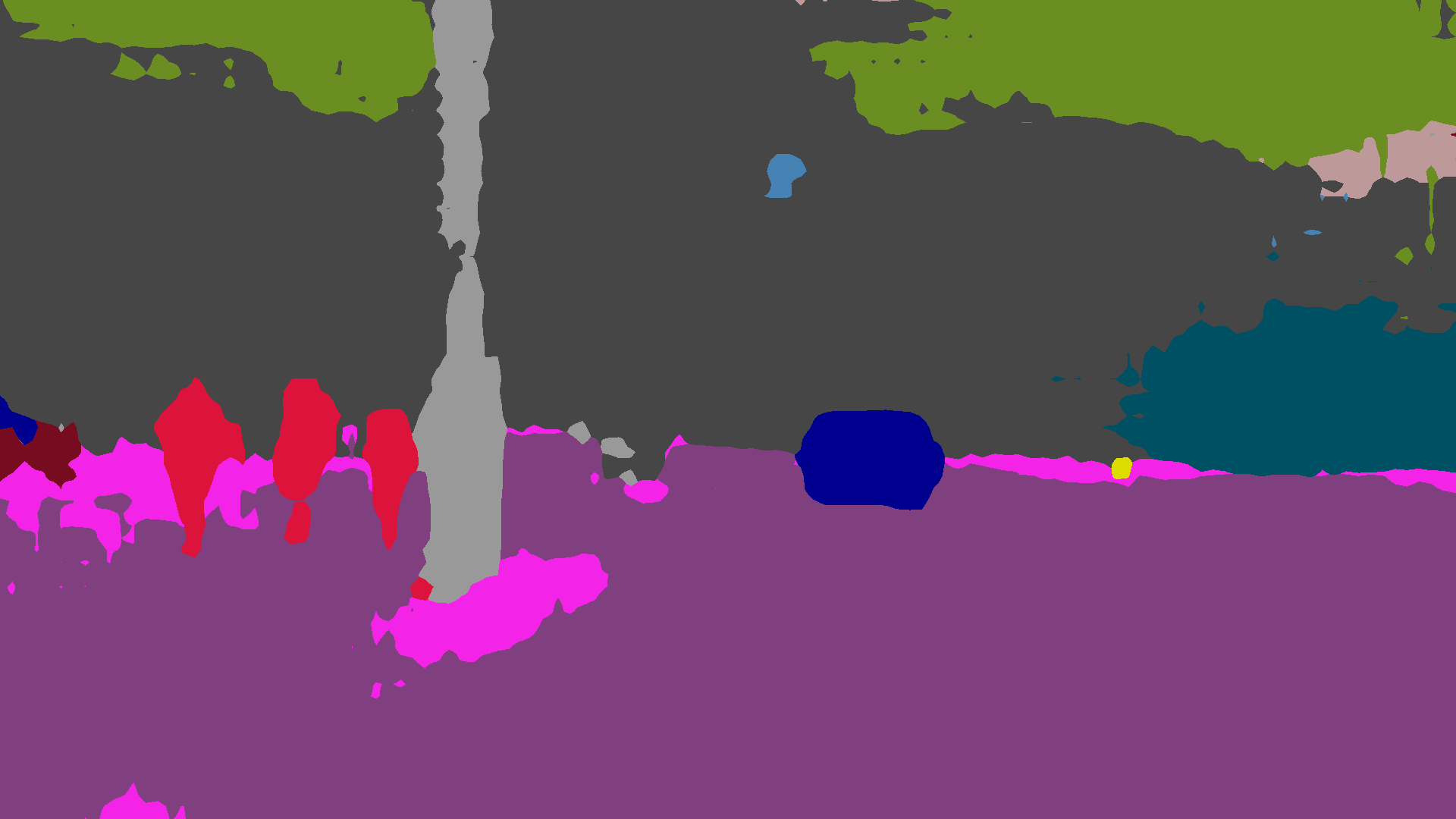} &
        \includegraphics[width=\linewidth,height=0.5625\linewidth]{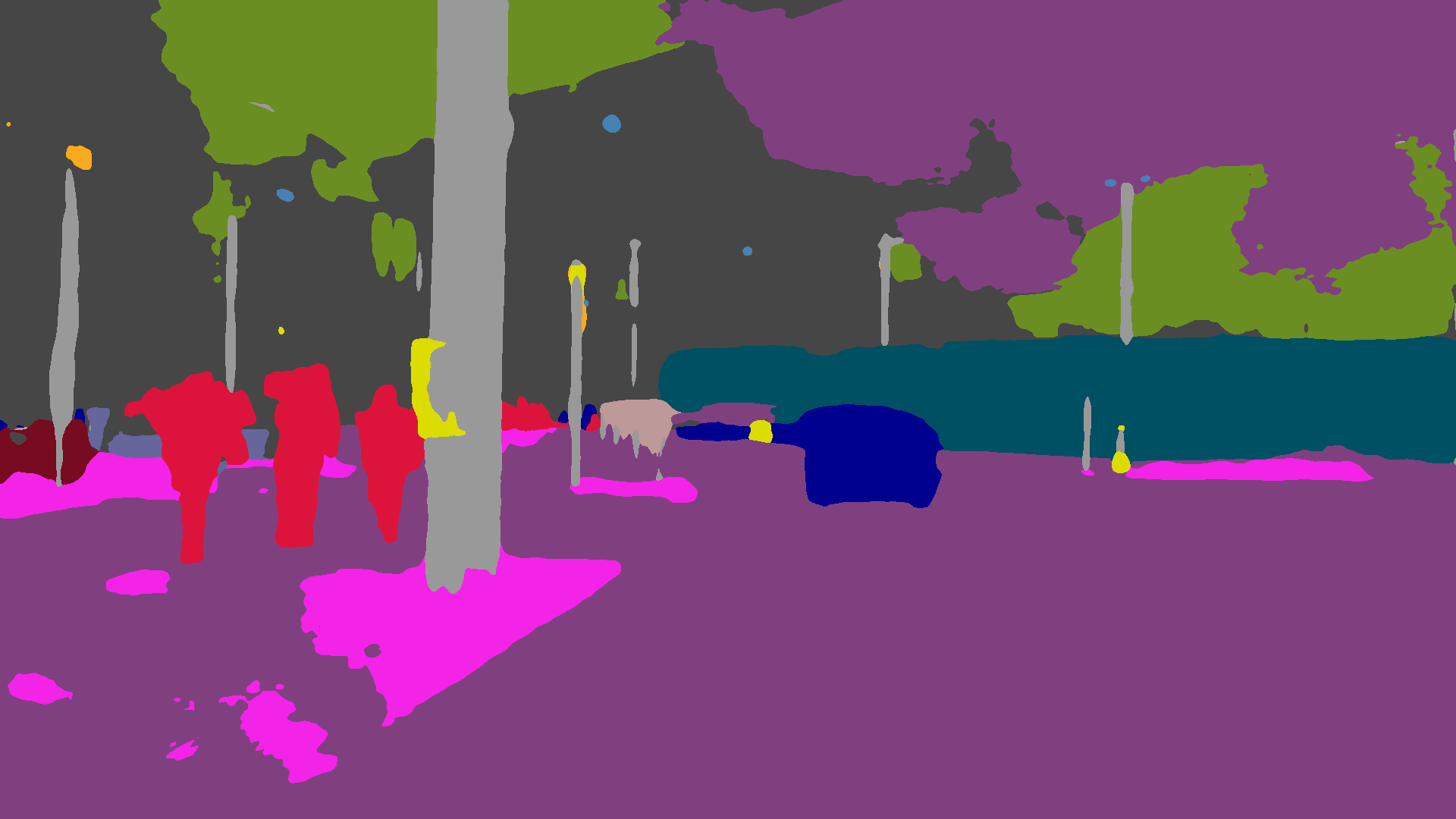} &
        \includegraphics[width=\linewidth,height=0.5625\linewidth]{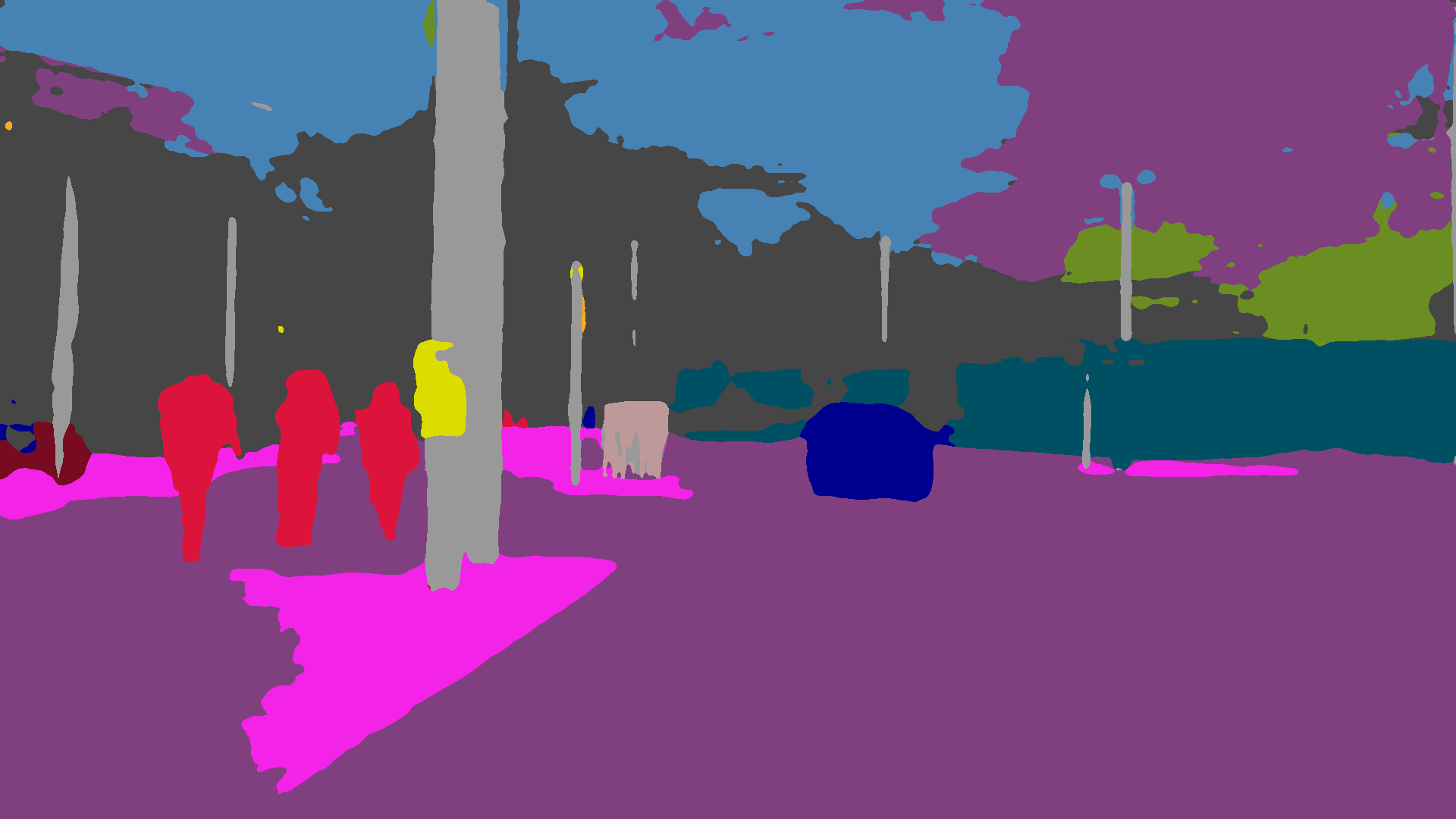} \\
        \includegraphics[width=\linewidth,height=0.5625\linewidth]{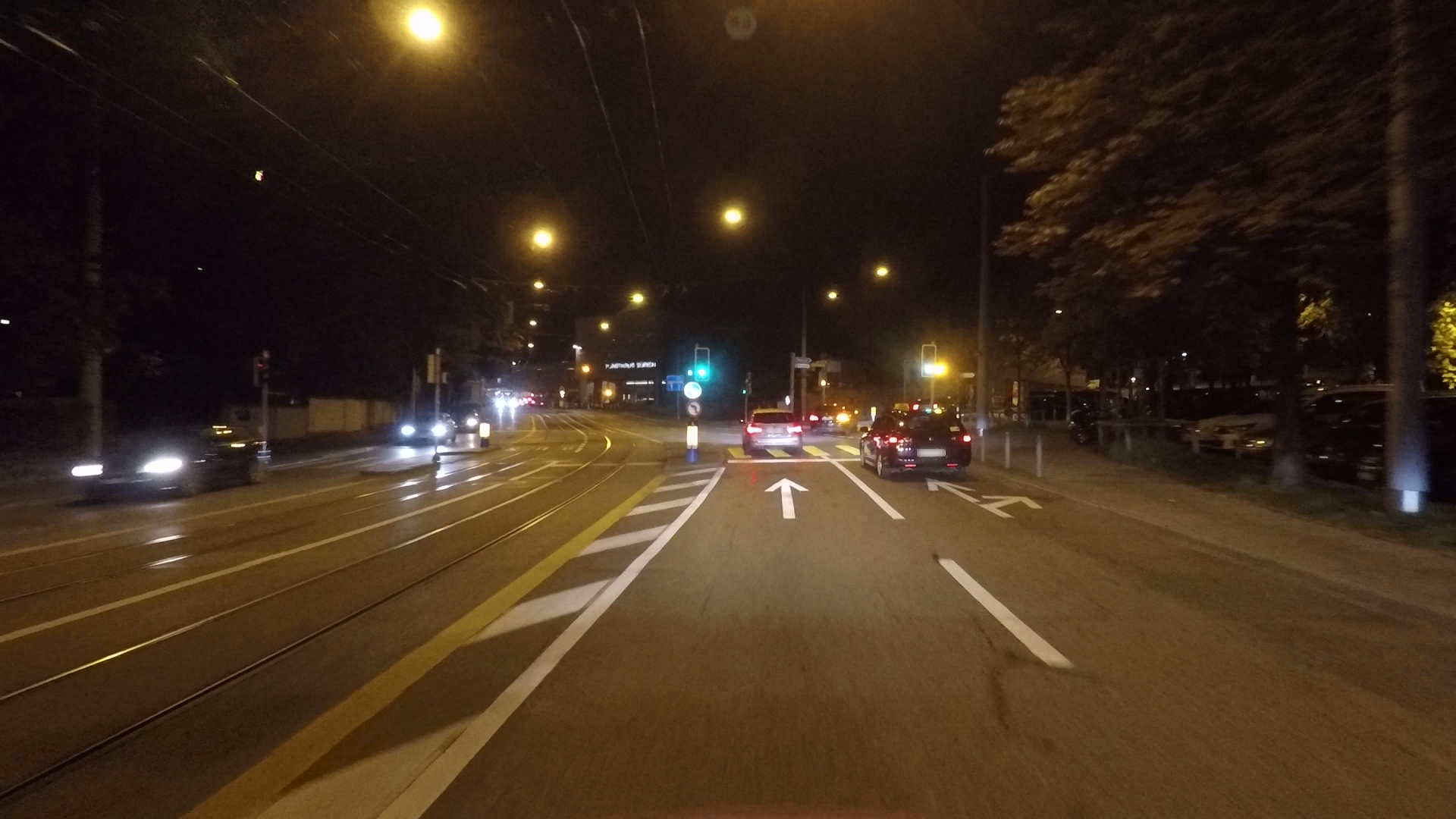} &
        \includegraphics[width=\linewidth,height=0.5625\linewidth]{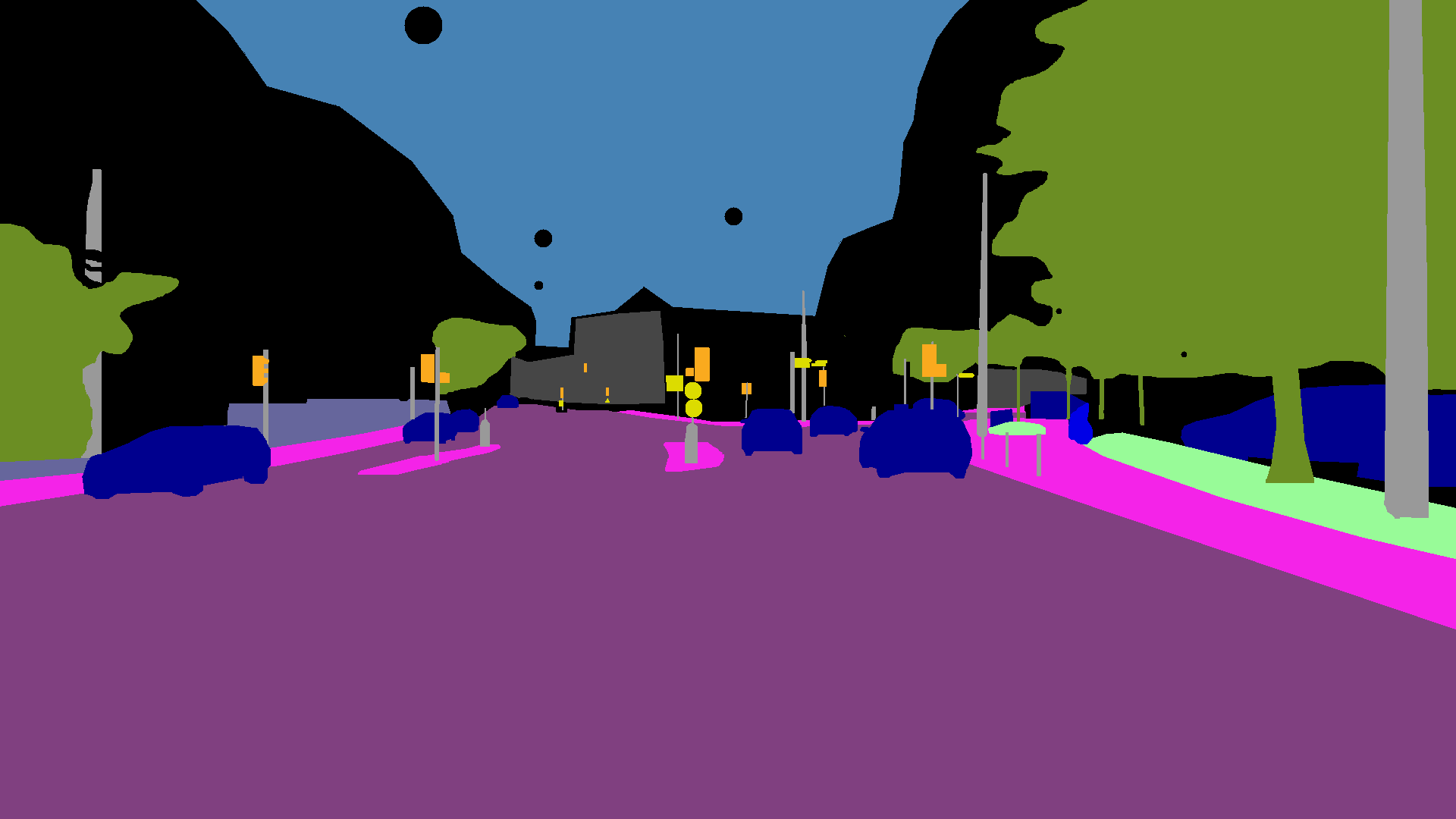} &
        \includegraphics[width=\linewidth,height=0.5625\linewidth]{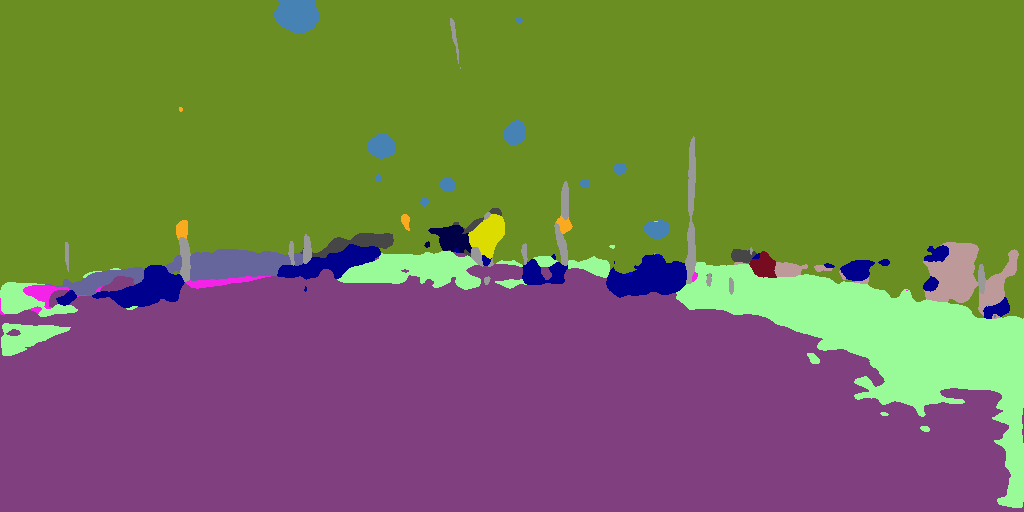} &
        \includegraphics[width=\linewidth,height=0.5625\linewidth]{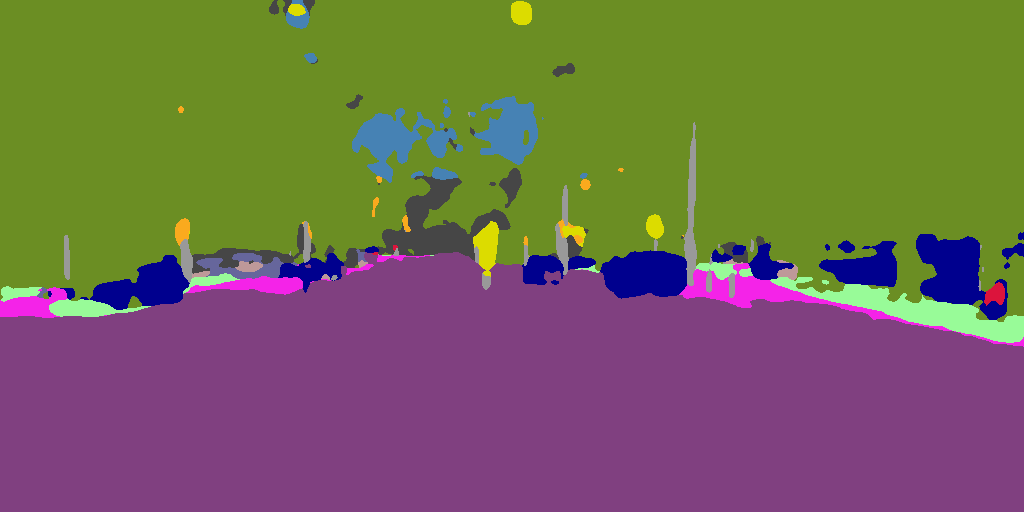} &
        \includegraphics[width=\linewidth,height=0.5625\linewidth]{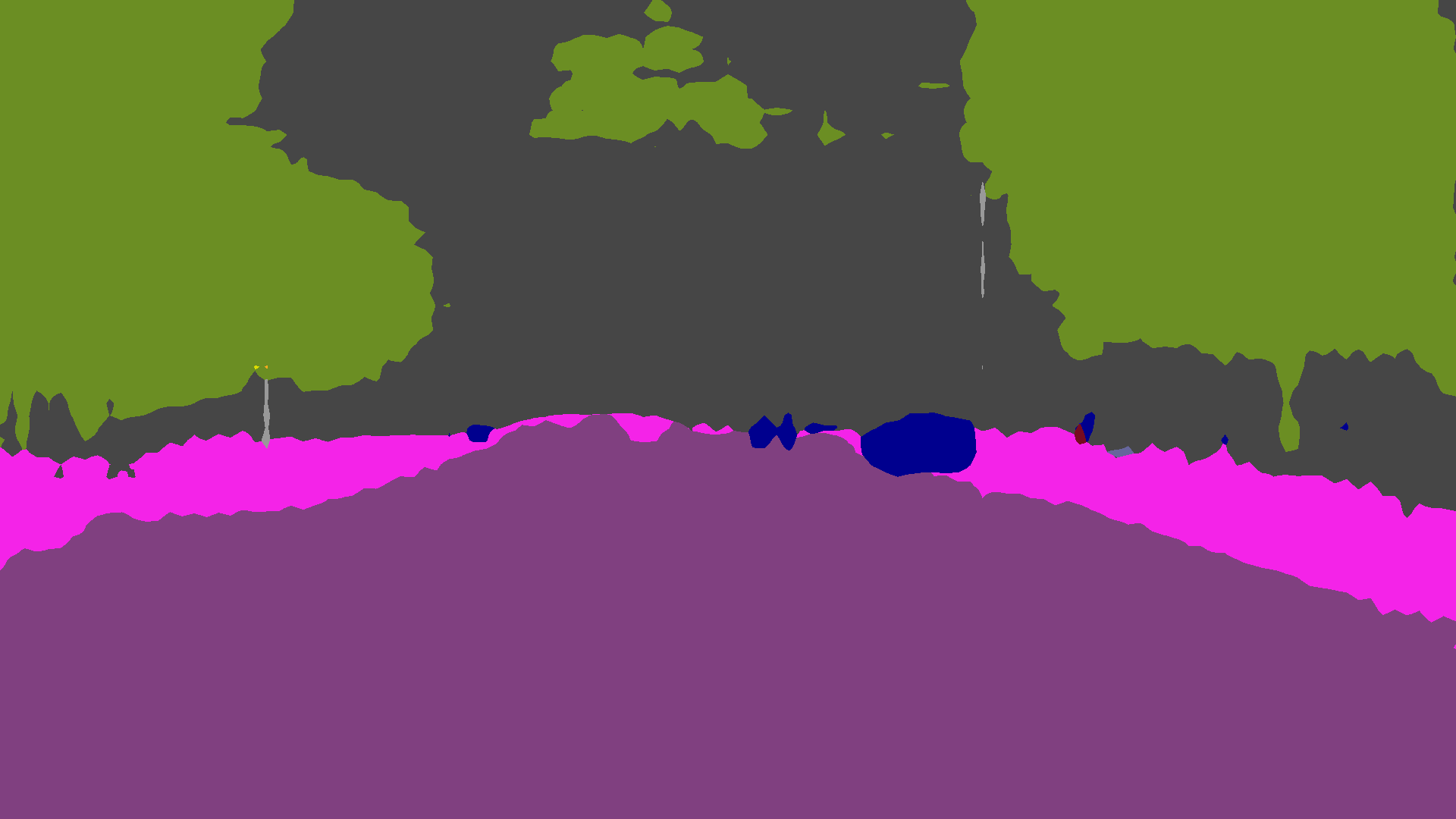} &
        \includegraphics[width=\linewidth,height=0.5625\linewidth]{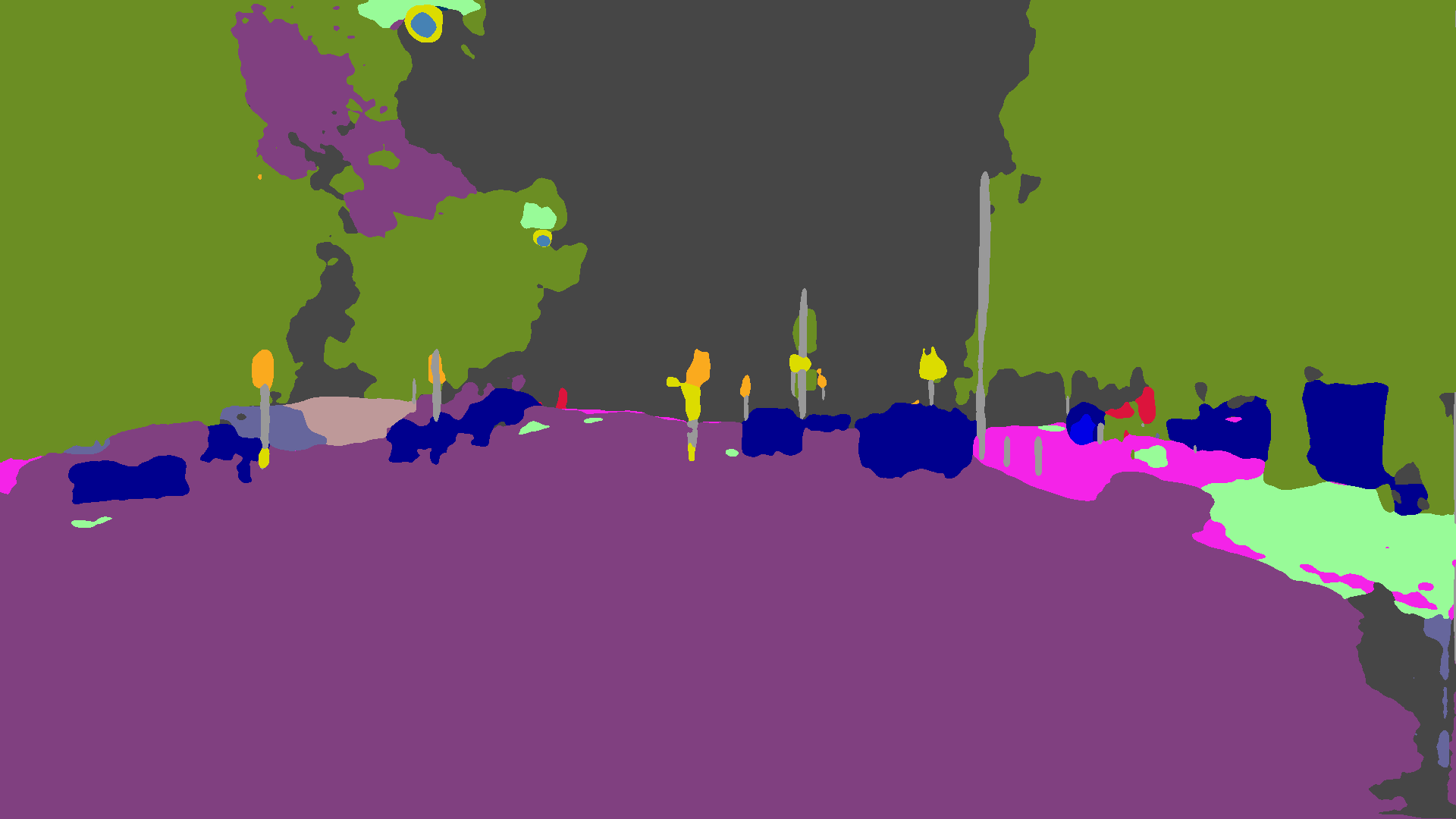} &
        \includegraphics[width=\linewidth,height=0.5625\linewidth]{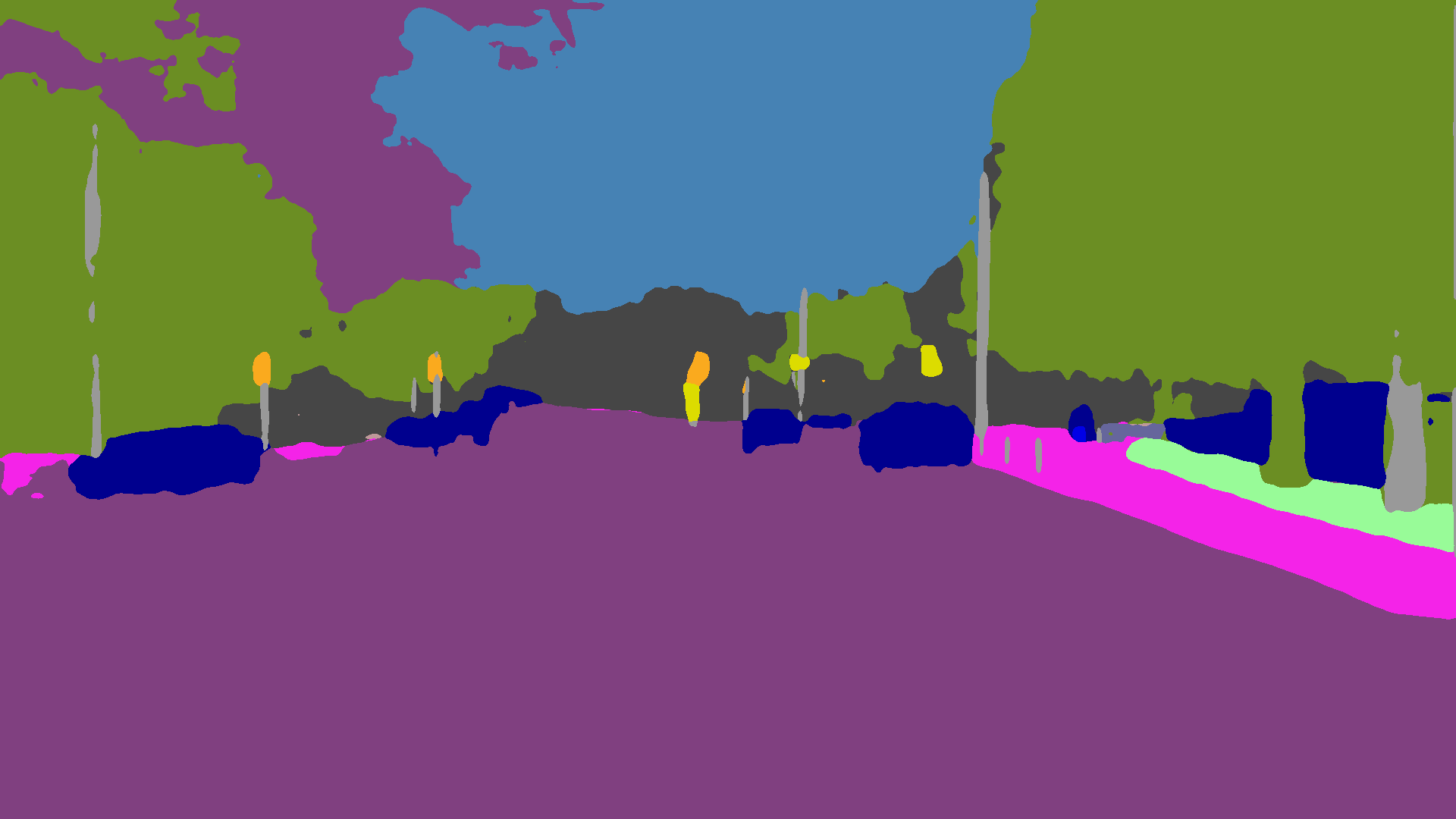} \\
        \vspace{5pt} Input &
        \vspace{5pt} GT & 
        \vspace{4pt} \makecell{RefineNet~\cite{Lin2017RefineNet} \\ (S)} & 
        \vspace{4pt} \makecell{$W$-RefineNet [ours] \\ (S)} & 
        \vspace{4pt} \makecell{AdaptSegNet~\cite{tsai2018learning} \\ (S+T)} & 
        \vspace{4pt} \makecell{DMAda~\cite{Dai18} \\ (S+T)} & 
        \vspace{4pt} \makecell{MGCDA~\cite{SDV20} \\ (S+T)}
    \end{tabularx}}
    \caption{Qualitative semantic segmentation results on the Dark Zurich~\cite{Sakaridis19} dataset. S and T indicate whether the model was trained on the source or target domain, respectively.}
    \label{fig:segmentation}
\end{figure*}



\subsection{Visual place recognition (VPR)}
We present results for VPR task in two phases: first, we compare against a similar work for place recognition based on a learnable normalisation of images~\cite{Jenicek2019NoFO}, and then we benchmark place representations based on color-invariant trained CNNs on an additional dataset, evaluation metric, and descriptor type to show broader applicability within VPR.

\textbf{Learnable normalisation.} We use the Tokyo 24/7 day-night place recognition dataset~\cite{Torii-CVPR2015} for this purpose, and follow the evaluation procedure described in~\cite{Jenicek2019NoFO}. To obtain place representations, the VGG Generalized Mean Pooling (GeM)~\cite{Radenovi2017FineTuningCI} network is prepended with our CIConv layer ($W$-VGG GeM) and trained on the Retrieval-SfM dataset as described in~\cite{Radenovi2017FineTuningCI}. The train dataset contains query images as well as both positive and negative target images of places photographed in daytime conditions. The results are reported as the mean Average Precision (mAP) in~\tab{placerec_results}. Results of competing methods are borrowed from Tables 1 and 2 in~\cite{Jenicek2019NoFO}. It can be observed that our method outperforms all models trained on daytime data only and achieves competitive results to the current state-of-the-art, which is an ensemble of two models trained on both daytime and nighttime data.

\begin{table}[]
\begin{tabularx}{\linewidth}{@{}l Y@{}}
\toprule
\textbf{Method} & \textbf{Tokyo 24/7 (mAP)} \\ \midrule
\multicolumn{1}{@{}l}{\textbf{Trained on source data only}} \\ \midrule
VGG GeM~\cite{Radenovi2017FineTuningCI} & 79.4 \\
$W$-VGG GeM [ours] & 83.3 \\
ResNet101 GeM~\cite{Radenovi2017FineTuningCI} & 85.0 \\
$W$-ResNet101 GeM [ours] & \textbf{88.3} \\
EdgeMAC~\cite{radenovic2018deep} & 75.9 \\
U-Net jointly~\cite{Jenicek2019NoFO} & 79.8 \\
CLAHE~\cite{zuiderveld94clahe} 	& 84.1 \\
EdgeMAC + VGG GeM~\cite{Jenicek2019NoFO} & 85.4 \\ \midrule
\multicolumn{1}{@{}l}{\textbf{Trained on source and target data}} \\ \midrule
VGG GeM~\cite{Radenovi2017FineTuningCI} & 79.8 \\
U-Net jointly~\cite{Jenicek2019NoFO} & 86.5 \\
CLAHE~\cite{zuiderveld94clahe} & 87.0 \\
EdgeMAC + CLAHE~\cite{Jenicek2019NoFO} & \textbf{90.5} \\
EdgeMAC + U-Net jointly~\cite{Jenicek2019NoFO} & 90.0 \\
\bottomrule
\end{tabularx}
\vspace*{-3mm}
\caption{Place recognition results on the Tokyo 24/7 dataset~\cite{Torii-CVPR2015}. VGG GeM with our CIConv layer outperforms all other methods trained on daytime data. + denotes an ensemble of different models.}
\label{tab:placerec_results}
\end{table}

\textbf{Broader VPR applicability.} Here, we use the two outdoor day-night datasets from VPRBench~\cite{zaffar2020vpr}: Gardens Point and Tokyo 24/7, where latter's evaluation is similar to the previous experiment but using Recall@1 as the evaluation metric in this case for both the datasets. For the Gardens Point dataset, we consider two settings: A (Appearance only) with only day-night variations and more challenging A+V (Appearance + Viewpoint) where viewpoint is also laterally shifted.  We consider three descriptor pooling types here using ImageNet-trained ResNet-101 (R101) as the backbone network: Maximum Activations of Convolutions (MAC)~\cite{tolias2015particular}, flattened tensor (Flat)~\cite{sunderhauf2015performance} and GeM, where only GeM is further trained on image retrieval task as described in the previous subsection. For all three descriptor types, we compute results for training with and without the prepended color invariant layer. Additionally, we compare against state-of-the-art VPR methods: DenseVLAD~\cite{torii201524} and AP-GeM~\cite{revaud2019learning}. 

In~\tab{placerec_results_2}, it can be observed that $W$-R101 GeM achieves state-of-the-art results for all datasets. Furthermore, all methods based on color invariant perform better than their vanilla counterparts, including the Flat and MAC descriptors. This shows that color invariant networks provide robust place representation for different pooling types even without VPR-specific training.

\begin{table}[]
\centering
\begin{tabularx}{\linewidth}{@{}lYYc@{}}
\toprule
\textbf{Method} & \textbf{GP:A+V} & \textbf{GP:A} & \textbf{Tokyo 24/7} \\
\midrule
AP-GeM~\cite{revaud2019learning}    & 0.87          & 0.92          & 0.91          \\
DenseVLAD~\cite{torii201524}    & 0.81          & 0.89          & 0.89          \\
R101 MAC~\cite{tolias2015particular}    & 0.51          & 0.56          & 0.20          \\
R101 Flat~\cite{sunderhauf2015performance}  & 0.56          & 0.68          & 0.84          \\
R101 GeM~\cite{Radenovi2017FineTuningCI}    & 0.90          & 0.96          & 0.91          \\
$W$-R101 MAC [ours]         & 0.53          & 0.70          & 0.20          \\
$W$-R101 Flat [ours]         & 0.61          & 0.91          & 0.85          \\
$W$-R101 GeM [ours]          & \textbf{0.94} & \textbf{0.97} & \textbf{0.93} \\

\bottomrule
\end{tabularx}
\vspace*{-3mm}
\caption{Recall@1 for VPR using different feature pooling types on Gardens Point (GP) and Tokyo 24/7 dataset. Color-invariant layer (W) based networks outperform their vanilla counterparts with W-R101-GeM achieving state-of-the-art results.}
\label{tab:placerec_results_2}
\end{table}
\section{Discussion}
The image formation model that lies at the foundation of the color invariants used in the CIConv layer is based on certain simplifying assumptions, such as purely matte reflections, non-transparent materials and a single, spatially uniform light source. Although most natural scenes do not satisfy these strict conditions, our results show that CNNs nevertheless do benefit from prior information derived from such approximate models. Moreover, current publicly available datasets, including the ones used in our experiments, are not appropriate for physics-based vision due to various artifacts introduced in post-processing steps (see Discussion in~\cite{Maxwell2019RealTimePR}). CIConv and other physics based methods can therefore only reach their full potential when sufficient attention is paid to preserving the physical correctness of the data during image capturing.

The robustness of color invariants to illumination changes comes at the loss of some discriminative power~\cite{Geusebroek2001Color}. The CIConv layer transforms the input image into an edge map representation that is no longer sensitive to the intensity and color of the light source, but as a side effect also removes valuable color information. We found that naively concatenating color invariants with the RGB input degrades performance, see section \textcolor{red}{3} of the supplementary material. Future research should therefore focus on implementing an adaptive mechanism for optimally combining color information and color invariant edge information. 

Zero-shot domain adaptation is a promising method for reducing the data dependency and the corresponding data collection and annotation costs in computer vision. We therefore hope that this paper inspires future research on integrating physics priors into neural networks.

\section*{Acknowledgements}
This project is supported in part by NWO (project VI.Vidi.192.100), the Australian Centre for Robotic Vision and the QUT Centre for Robotics.

{\small
\bibliographystyle{ieee_fullname}
\bibliography{egbib}
}

\end{document}